\documentclass[sigconf]{acmart}
\AtBeginDocument{%
  }

\copyrightyear{2025}
\acmYear{2025}
\setcopyright{cc}
\setcctype{by}
\acmConference[MM '25] {Proceedings of the 33rd ACM International Conference on Multimedia}{October 27--31, 2025}{Dublin, Ireland.}
\acmBooktitle{Proceedings of the 33rd ACM International Conference on Multimedia (MM '25), October 27--31, 2025, Dublin, Ireland}
\acmISBN{979-8-4007-2035-2/2025/10}
\acmDOI{10.1145/3746027.3754822}

\usepackage{subcaption}
\usepackage{amsfonts}
\usepackage[inkscapelatex=false]{svg}
\usepackage[linesnumbered,ruled]{algorithm2e}

\newcommand{\G}{{\mathcal{G}}}
\newcommand{\V}{{\mathcal{V}}}

\newcommand{\LL}{{\mathcal{L}}}

\newcommand{\D}{{\mathcal{D}}}
\newcommand{\M}{{\mathcal{M}}}
\newcommand{\T}{{\mathcal{T}}}

\begin{document}

\title{Towards Effective Open-set Graph Class-incremental Learning}

\author{Jiazhen Chen}
\orcid{0000-0001-9962-2974}
\orcid{1234-5678-9012}
\affiliation{%
  \institution{University of Waterloo}
  \city{Waterloo}
  \state{Ontario}
  \country{Canada}
}
\email{j385chen@uwaterloo.ca}

\author{Zheng Ma}
\orcid{0009-0005-6449-5983}
\affiliation{%
  \institution{University of Waterloo}
  \city{Waterloo}
    \state{Ontario}
  \country{Canada}}
\email{z43ma@uwaterloo.ca}

\author{Sichao Fu}
\authornote{Corresponding author}
\orcid{0000-0002-4363-1000}
\affiliation{%
  \institution{Huazhong University of Science and Technology}
  \city{Wuhan}
  \state{Hubei}
  \country{China}
}
\email{fusichao_upc@163.com}

\author{Mingbin Feng}
\orcid{0000-0002-9748-6435}
\affiliation{%
  \institution{University of Waterloo}
  \city{Waterloo}
    \state{Ontario}
  \country{Canada}}
\email{ben.feng@uwaterloo.ca}

\author{Tony S. Wirjanto}
\orcid{0000-0003-1324-9131}
\affiliation{%
  \institution{University of Waterloo}
  \city{Waterloo}
    \state{Ontario}
  \country{Canada}}
\email{twirjanto@uwaterloo.ca}

\author{Weihua Ou}
\authornotemark[1]
\orcid{0000-0001-5241-7703}
\affiliation{%
  \institution{Guizhou Normal University}
  \city{Guiyang}
  \state{Guizhou}
  \country{China}}
\email{ouweihua@gznu.edu.cn}

\renewcommand{\shortauthors}{Chen et al.}

\begin{abstract}
Graphs play a pivotal role in multimedia applications by integrating information to model complex relationships. Recently, graph class-incremental learning (GCIL) has garnered attention, allowing graph neural networks (GNNs) to adapt to evolving graph analytical tasks by incrementally learning new class knowledge while retaining knowledge of old classes.  
Existing GCIL methods primarily focus on a closed-set assumption, where all test samples are presumed to belong to previously known classes. Such assumption restricts their applicability in real-world scenarios, where unknown classes naturally emerge during inference, and are absent during training. 
In this paper, we explore a more challenging open-set graph class-incremental learning scenario with two intertwined challenges: catastrophic forgetting of old classes, which impairs the detection of unknown classes, and inadequate open-set recognition, which destabilizes the retention of learned knowledge.
To address the above problems, a novel OGCIL framework is proposed, which utilizes pseudo-sample embedding generation to effectively mitigate catastrophic forgetting and enable robust detection of unknown classes. To be specific, a prototypical conditional variational autoencoder is designed to synthesize node embeddings for old classes, enabling knowledge replay without storing raw graph data. To handle unknown classes, we employ a mixing-based strategy to generate out-of-distribution (OOD) samples from pseudo in-distribution and current node embeddings. A novel prototypical hypersphere classification loss is further proposed, which anchors in-distribution embeddings to their respective class prototypes, while repelling OOD embeddings away. Instead of assigning all unknown samples into one cluster, our proposed objective function explicitly models them as outliers through prototype-aware rejection regions, ensuring a robust open-set recognition. Extensive experiments on five benchmarks demonstrate the effectiveness of OGCIL over existing GCIL and open-set GNN methods.

\end{abstract}




\maketitle

\section{Introduction}

In recent years, graphs have played a pivotal role in multimedia applications such as social networks, recommendation systems, and content-based retrieval, by integrating diverse data and modeling intricate interactions~\cite{wang2023tiva,li2019routing,pang2024ftf,yoshida2015heterogeneous}. 
To derive meaningful representations from graphs, graph neural networks (GNNs)~\cite{kipf2017semi} have emerged as a powerful tool, leveraging both node attributes and graph topology for tasks such as node classification and link prediction. Despite their success, GNNs typically assume graphs exist with consistent training and test distributions. Nevertheless, real-world graphs evolve continually, introducing new nodes, edges, and even novel classes over time. For instance, social networks expand as users join and form connections, while recommendation systems must accommodate newly introduced items and user interactions~\cite{he2021dynamic,chen2022recommendation}. While retraining on each graph update could adapt to such changes, it is impractical owing to significant memory, computational, and privacy limitations~\cite{ren2023incremental}. Conversely, focusing solely on new data without revisiting past information risks catastrophic forgetting, where the model fails to retain knowledge of previously learned classes.

Graph class-incremental learning (GCIL) has recently emerged as a key research focus, which requires the GNNs to incrementally learn new tasks with disjoint classes on emerging subgraphs without full access to past classes~\cite{febrinanto2023graph}. To tackle this problem, an effective GCIL method must maintain knowledge of earlier classes to prevent catastrophic forgetting, while effectively learning newly emerged classes. To this end, numerous GCIL methods have been proposed. For instance, replay-based approaches store or generate pseudo-samples to approximate the data distribution of older classes~\cite{streamingGNN,ergnn,sgnn-gr,replayGNN}. Regularization-based methods constrain parameter updates to preserve previously learned knowledge~\cite{twp,geometer,ssrm}. Whereas representation-based techniques aim to maintain task-invariant embeddings and refine task-specific components~\cite{tpp,hpn,dicgrl}. While GCIL methods handle incremental tasks well, they seriously rely on the closed-set assumption that all test samples belong to classes learned so far in the training process. However, the open-set scenarios, where samples from novel classes appear during inference, are frequently observed in real-world applications. For instance, social networks may see new user groups~\cite{ren2022known}, and recommendation systems often face unseen item categories~\cite{mahdavi2021survey}. Without robust mechanisms to handle truly unknown categories, GCIL methods are prone to misclassify them as known classes due to overconfidence in their closed-set assumptions.

Recently, open-set recognition (OSR) has gained increasing attention in the graph domain, which aims at identifying and rejecting inputs from unknown classes while maintaining accurate classification for known classes. Existing graph OSR methods employ various strategies, such as increasing entropy to enhance the detection of unknown samples~\cite{openWGL}, generating pseudo-unknown samples to simulate unseen classes~\cite{zhang2023g2pxy}, and leveraging neighbor information to refine the distinction between known and unknown nodes~\cite{OSSNC,openWRF}. However, these proposed approaches presuppose access to training data from all known classes, and their effectiveness is strongly tied to the accurate representation of these classes. To the best of our knowledge, no existing works attempt to jointly tackle GCIL and OSR under a unified setting in the graph domain. Integrating these two tasks is non-trivial owing to their inherent interdependence. Specifically, inadequate retention of knowledge from older classes can result in forgotten classes being mislabeled as unknown, while poor handling of unknown classes can destabilize the knowledge learned for previously seen classes. 

In this study, we propose a new framework to address both the catastrophic forgetting problem and unknown class detection under the unified GCIL and OSR setting, termed OGCIL. The core idea of OGCIL is to generate pseudo samples to compensate for the lack of historical and unknown class data. However, due to the interdependencies and structural complexity of graphs, generating raw graph data is inherently challenging in comparison to raw data. As a compromise, OGCIL operates at the embedding level to simplify the generation process while preserving essential graph information. Specifically, we decouple the learning of node features into two parts: less-task-invariant embeddings, which are regularized by knowledge distillation to remain stable across tasks, and a task-variant encoder, incrementally tuned to quickly adapt to distributional changes. A prototype-based loss is further employed to cluster embeddings around class-specific prototypes, and a conditional variational autoencoder (CVAE) enforces a normal distribution in the latent space (i.e., the encoded output) centered on these prototypes. Subsequently, we can facilitate the generation of diverse pseudo in-distribution (ID) embeddings via the CVAE, and out-of-distribution (OOD) samples can then be generated by mixing pseudo ID samples with new class and pseudo ID embeddings.

One key challenge in OSR is the ambiguity of OOD samples, particularly those generated through a mixing strategy, as they may lie between the embedding spaces of multiple known classes. Similarly, actual unknown samples may span several unknown classes, making it difficult to cluster them under a single category. To address this, OGCIL introduces a novel prototypical hypersphere classification loss, which establishes context-aware decision boundaries by anchoring ID samples near their respective class prototypes while rejecting all unknowns and samples from other classes as outliers. By avoiding the restrictive assumption of grouping all unknowns into a single prototype, we enable more robust handling of diverse unknown classes and noisy pseudo samples.

To summarize, our contributions are listed as follows:
\begin{itemize}

    \item \textbf{Novel Problem:} We make the first attempt to address the dual challenges of catastrophic forgetting and unknown class detection in a unified GCIL and OSR setting on graphs. To our knowledge, this problem has not been explored before.
   \item \textbf{Pseudo Sample Generation:} To address the data scarcity for historical and unknown classes, 
   a prototypical CVAE is proposed to synthesize pseudo ID samples directly in the embedding space, subsequently OOD samples are obtained through interpolation of pseudo and current embeddings.
    \item \textbf{Prototypical Hypersphere Classification Loss:} To address the complexities posed by noisy pseudo-samples and unknown inputs, we propose a prototypical hypersphere classification loss that encloses ID samples within class-specific hyperspheres, while treating pseudo OOD and off-class samples as outliers.
    \item \textbf{Extensive Experimental Validation:} We conduct comprehensive experiments on five real-world graph datasets under the class-incremental open-set setting. Extensive experiments show the superior performance of the proposed OGCIL framework over a variety of benchmark methods.
\end{itemize}

\section{Related Work}
\subsection{Graph Class-incremental Learning}

Graph incremental learning enables GNNs to sequentially learn new knowledge on evolving graph-related tasks with disjoint classes~\cite{febrinanto2023graph,CIL_survey}. Two prominent variants include graph task-incremental learning, which uses task-specific classifiers, and graph class-incremental learning (GCIL), which requires classification across all learned classes without task identifiers. Thus, GCIL is often considered a more practical and challenging setting. Existing GCIL approaches fall into three categories: replay-based~\cite{streamingGNN,ergnn,sgnn-gr,replayGNN}, regularization-based~\cite{twp,geometer,ssrm}, and representation-based approaches~\cite{tpp,hpn,dicgrl}.


Replay-based methods focus on replaying historical graph data or generating pseudo samples from previous classes. StreamingGNN~\cite{streamingGNN} maintains a memory buffer of historical graph data using random sampling, while ER-GNN~\cite{ergnn} selects representative nodes for old classes via mean-based and coverage maximization sampling. 
Instead of storing real graph data, SGNN-GR~\cite{sgnn-gr} uses a GAN-based framework to generate pseudo-node sequences via random walks, training a GNN on both real and pseudo samples. In contrast, our framework generates pseudo-samples directly in the embedding space and is compatible with existing sampling-based methods.
\begin{figure*}[t]
\centering
\includegraphics[width=\linewidth]{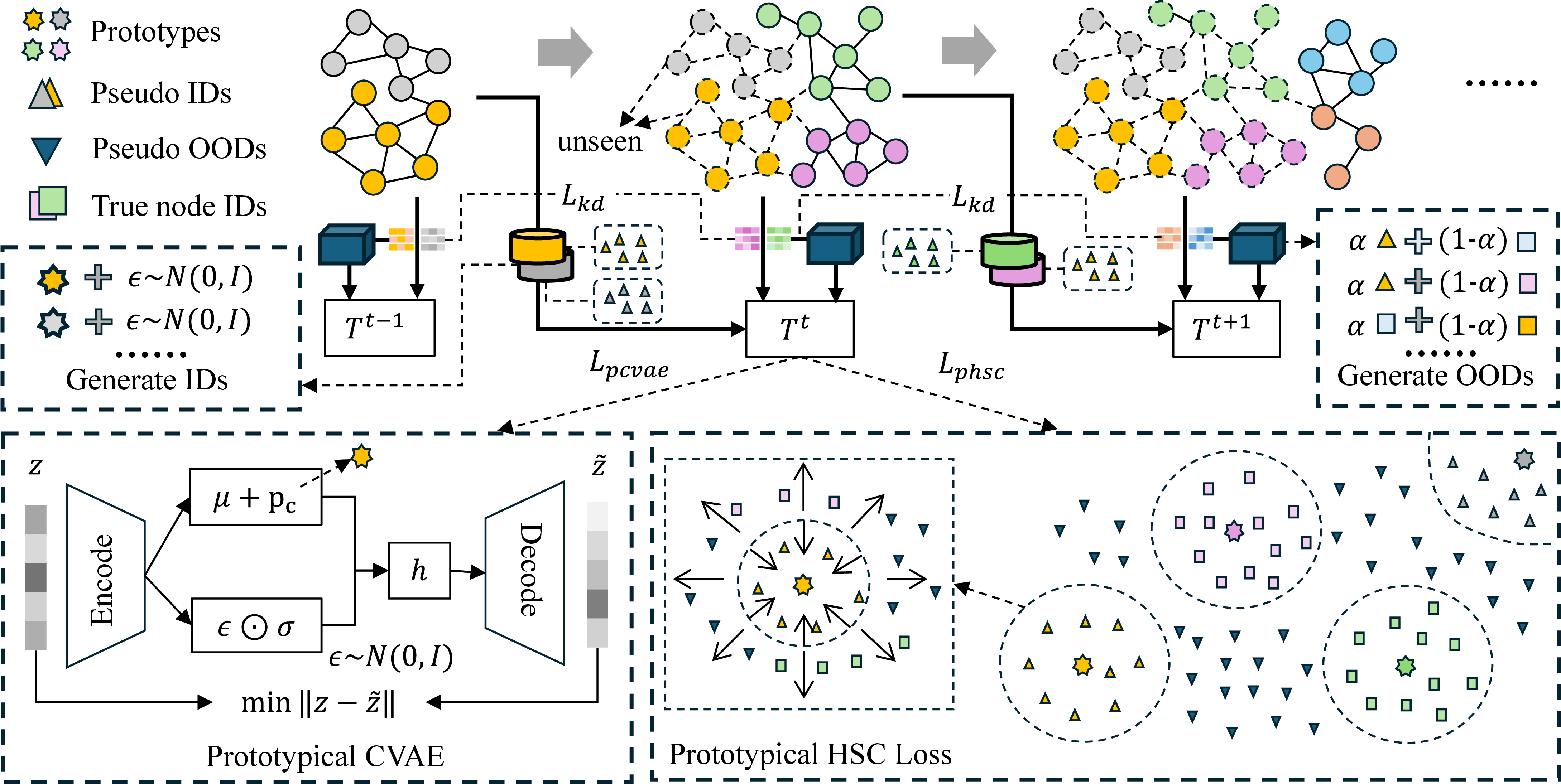}
\caption{A diagram illustrating the proposed OGCIL framework.}
\label{fig: framework}
\end{figure*}

Regularization-based methods constrain parameter updates to retain knowledge from previous classes while learning new ones. TWP~\cite{twp} encodes topological information using attention mechanisms and applies gradient-based regularization to retain parameters critical to previously learned classes. Geometer~\cite{geometer} employs knowledge distillation, which transfers softened logits from a teacher model to a student model to preserve class boundaries across tasks. SSRM~\cite{ssrm} aligns embeddings across tasks by incorporating maximum mean discrepancy-based regularization. Similarly, our method employs knowledge distillation for embedding stability, though this is modular and compatible with alternative regularizations.

Representation-based methods separate task-invariant from task-variant embeddings to reduce interference. DiCGRL~\cite{dicgrl} disentangles graph embeddings into semantic components and updates only the task-variant parts. TPP~\cite{tpp} combines Laplacian smoothing for task profiling with task-specific graph prompts and leverages a frozen pre-trained GNN for task-variant representation learning.  Analogously, our approach uses a VAE-based prompt module to capture task-specific variations, while stabilizing the primary GNN with distillation-based regularization.


\subsection{Open-set Recognition in Graphs}
Open-set recognition (OSR) detects inputs from unseen classes at test time while preserving accuracy on known classes. 
Existing OSR methods primarily focus on domains such as image and text data, employing approaches like confidence-based adjustments~\cite{odin,openMAX,objectobsphere}, distance metrics~\cite{isomax,crosr,snell2017prototypical}, and generative modeling~\cite{g-openmax,cdgl,chen2021adversarial}. 
For instance, ODIN~\cite{odin} boosts unknown detection via temperature scaling and input perturbations, IsoMax~\cite{old_isomax} separates samples by their distance to isotropic prototypes, and G-OpenMax~\cite{g-openmax} uses GANs to synthesize unknowns for greater robustness.

Recent works have started to adapt OSR to the graph domain. OpenWGL~\cite{openWGL} employs variational embeddings and a class uncertainty loss to increase entropy for unknown nodes, thus detecting them via entropy-based sampling. OSSNC~\cite{OSSNC} adjusts neighbor influence during message passing to minimize mixing between ID and OOD nodes, with bi-level optimization to reduce overfitting. OpenIMA~\cite{OpenIMA} addresses variance imbalance between seen and novel classes by generating high-confidence pseudo-labels through clustering and aligning them with class labels. G$^2$Pxy~\cite{zhang2023g2pxy} generates proxy unknown nodes using manifold mixup and integrates these proxies into an open-set classifier that treats unknowns as a single category. However, these methods lack class-incremental capabilities, leaving them vulnerable to catastrophic forgetting when old-class data becomes inaccessible during incremental updates. 

OpenWRF~\cite{openWRF} and LifeLongGNN~\cite{Galke2021Lifelong} target lifelong learning in open-world graph settings but differ fundamentally from our setup. OpenWRF assumes full access to historical data and focuses on OSR via OOD score propagation and a lightweight classifier. LifeLongGNN adapts incrementally using warm restarts and expands its classifier for classes only observed during training, assuming same class distributions in training and testing. Neither method is designed to tackle the dual challenges of knowledge forgetting and unknown class detection.

\section{Methodology}
\subsection{Problem Formulation}

We study a setting in which a model is trained on a sequence of tasks with disjoint classes. In this setting, the model must accurately classify nodes from classes seen up to the current task while rejecting nodes from classes that were never encountered during training. Specifically, we address two key challenges: 1) catastrophic forgetting—retaining previous knowledge when new classes are introduced, and 2) open-set recognition—accurately detecting nodes from unseen classes during inference.


Formally, we consider a sequence of tasks $\mathcal{T}=\{\mathcal{T}^1, \mathcal{T}^2, \ldots, \mathcal{T}^M\}$. Each task $\mathcal{T}^t$ is associated with a node classification objective on a newly emerged attributed subgraph $\mathcal{G}^t = (\mathcal{V}^t, X^t, A^t)$, where $\mathcal{V}^t$ denotes the node set, $X^t$ denotes node features, and $A^t$ denotes the adjacency matrix capturing edge information. Nodes in each task are partitioned into a training set $\mathcal{V}^t_{\mathrm{tr}}$ and a test set $\mathcal{V}^t_{\mathrm{te}}$.

In each training stage of task $\mathcal{T}^t$, the training set $\mathcal{V}^t_{\mathrm{tr}}$ consists of labeled nodes belonging to newly introduced classes $\mathcal{Y}^t_{\mathrm{tr}}$, which are disjoint from classes introduced in previous tasks, i.e., 
$
  Y^t_{\mathrm{tr}} \,\cap\, \Bigl(\bigcup_{\tau=1}^{t-1} Y^{\tau}_{\mathrm{tr}}\Bigr) = \emptyset.
$
To alleviate catastrophic forgetting, the model can leverage a small exemplar set $\mathcal{M}^t$ (a subset of historical nodes, if available) during training for each task.
During inference, the test set $\mathcal{V}^t_{\mathrm{te}}$ includes nodes from two categories: 1) Known classes ($\mathcal{V}^t_{\mathrm{te},k}$): Nodes belonging to classes introduced in the task $\mathcal{T}^t$. 2) Unknown classes ($\mathcal{V}^t_{\mathrm{te},\mathrm{unk}}$): Nodes from classes that have never been observed during training.
We adopt an inductive setting, where the model does not observe unknown nodes in training for each task. However, for data efficiency and practical realism, we assume any class designated as unknown in task $\mathcal{T}^{t-1}$ becomes known in the subsequent task $\mathcal{T}^t$. Note that under the inductive setting, these newly labeled classes are effectively ``new'' since their nodes were never observed or labeled in early tasks.


The overall goal of OGCIL is thus twofold: 1) Accurate classification: Correctly classify nodes from any previously or currently learned classes $\bigcup_{\tau=1}^{t} \mathcal{V}^{\tau}_{\mathrm{te},k}$. 2) Open-set recognition: nodes from the newly emerged unseen classes $\mathcal{V}_{\mathrm{te},\mathrm{unk}}^{t}$ should be recognized as ``unknown'', implying the model chooses from $C+1$ outcomes (i.e., $C$ known classes + 1 unknown label).

\subsection{Overview}

The cornerstone of our framework lies in the generation of high-quality pseudo samples for both known classes (in-distribution samples) and unknown classes (out-of-distribution samples), addressing catastrophic forgetting while facilitating robust detection of unknown classes. For ID sample generation, we propose a prototypical CVAE that reconstructs and samples node embeddings for previously learned classes, eliminating the need for raw graph structures (Sections~\ref{class-conditional VAE} and~\ref{pseudo id sample generation}). Additionally, we incorporate knowledge distillation to ensure the stability of GNNs output representations across tasks (Section~\ref{knowledge distillation}). To identify emerging unknown classes in new tasks, we synthesize OOD samples by mixing embeddings from different known classes and pseudo ID embeddings of historical classes (Section~\ref{mixing strategy for ood}). As these mixed embeddings inherently deviate from any known class prototype, clustering them into a single “unknown” class is suboptimal. To address this, we introduce a prototypical hypersphere classification loss, which constructs context-aware hypersphere boundaries around class-specific embeddings, while treating embeddings from other classes and mixed OOD samples as outliers and rejecting them from the hypersphere (Section~\ref{prototype HSC}). Section~\ref{overall_training} shows the overall training objective.

\subsection{In-distribution Sample Generation}

In class-incremental learning, the lack of historical class samples can cause GNNs to overfit to newly introduced classes. Given the high dimensionality and complexity of reconstructing raw graph data, we instead generate pseudo-nodes directly in the embedding space, assuming embeddings effectively capture essential graph structures. To mitigate the potential representation drift across incremental tasks, we decouple the learning of node representations into two synergistic components as inspired by recent advances in graph prompt tuning~\cite{sun2023graph}: a task-invariant representation stabilized via knowledge distillation (akin to a pre-trained GNN), and a lightweight encoder that dynamically ``prompt'' to structural and distributional variations in incremental tasks. We then introduce a prototypical CVAE to generate pseudo node embeddings by variationally sampling from a latent probabilistic distribution parameterized by class prototypes.


\subsubsection{Prototypical Conditional Variational Autoencoder}
\label{class-conditional VAE}

Concretely, we consider a CVAE in which each class $c$ is associated with a learnable prototype $\mathbf{p}_c$. Let $\mathbf{z} \in \mathbb{R}^d$ denote the GNN-generated embedding of a node. The VAE encoder $q_{\phi}(\mathbf{h} \mid \mathbf{z})$ defines a posterior distribution over the latent variable $\mathbf{h}$. For a training sample $\mathbf{z}$ of class $c$, we impose a Kullback-Leibler (KL) penalty to align $q_{\phi}(\mathbf{h} \mid \mathbf{z})$ with the Gaussian prior $\mathcal{N}(\mathbf{p}_c, \mathbf{I})$. Meanwhile, the decoder $p_{\theta}(\mathbf{z} \mid \mathbf{h})$ reconstructs $\mathbf{z}$ from $\mathbf{h}$.
Formally, given a task $\T^{t}$, the prototypical CVAE loss for a sample $\mathbf{z}$ belonging to class $c$ is:
\begin{align}
    \mathcal{L}_{\mathrm{pcvae}}^t
    &=  \frac{1}{|\D^t|} \sum_{\mathbf{z} \in 
    \D^t
    } \Bigl[
    -\lambda_{\mathrm{reconst}} \, \mathbb{E}_{q_{\phi}(\mathbf{h}\mid \mathbf{z})}\bigl[\log p_{\theta}(\mathbf{z}\mid \mathbf{h})\bigr] 
    \notag \\
    &+ \mathrm{KL}\bigl(q_{\phi}(\mathbf{h}\mid \mathbf{z}) \,\|\, \mathcal{N}(\mathbf{p}_c, \mathbf{I})\bigr)  \Bigr].  \label{eq.pcvae} 
\end{align}
Here, $\D^t=\mathcal{D}^t_{\mathrm{real}} \cup \mathcal{D}^{t-1}_{\mathrm{ID}}$, where $\mathcal{D}^t_{\mathrm{real}}$ denotes the set of GNN embeddings for nodes in $\V^t_{\mathrm{tr}}$, and $\mathcal{D}^{t-1}_{\mathrm{ID}}$ denotes the set of pseudo ID embeddings generated from the previous task $\T^{t-1}$. The expectation $\mathbb{E}_{q_{\phi}(\mathbf{h}\mid \mathbf{z})}[\cdot]$ is approximated by sampling $\mathbf{h}$ via the reparameterization trick:
\begin{align}
    \mathbf{h} \;=\; \mu_{\phi}(\mathbf{z}) \;+\; \sigma_{\phi}(\mathbf{z}) \,\odot\, \boldsymbol{\epsilon}, 
    \quad 
    \boldsymbol{\epsilon} \,\sim\, \mathcal{N}\bigl(\mathbf{0}, \mathbf{I}\bigr).
\end{align}
Here, $\mu_{\phi}(\mathbf{z})$ and $\sigma_{\phi}(\mathbf{z})$ are outputs of the encoder network $q_{\phi}(\mathbf{h}\mid \mathbf{z})$, which parameterizes the approximate posterior distribution $\mathcal{N}\!\bigl(\mu_{\phi}(\mathbf{z}),\,\mathrm{diag}(\sigma_{\phi}(\mathbf{z})^2)\bigr)$.
In parallel, we jointly learn each prototype $\mathbf{p}_c$ through a distance-based classification loss that encourages GNN embeddings from class $c$ to cluster around $\mathbf{p}_c$ while remaining distinct from other class prototypes (detailed in Section~\ref{prototype HSC}). 

Note that neither the CVAE encoder nor the decoder explicitly takes class $c$ as input; the class conditionality arises from the latent prior centered on $\mathbf{p}_c$. Compared to traditional  CVAEs~\cite{sohn2015learning} that use one-hot encoded class labels, $\mathbf{p}_c$ provides a richer and more expressive representation, inherently capturing intra-class cohesion and inter-class relationships. Furthermore, Eq.~\ref{eq.pcvae} remains consistent with the classification loss, which encloses samples in a hypersphere around $\mathbf{p}_c$. Incorporating $\mathbf{p}_c$ into the KL prior ensures that pseudo samples align with the same center, harmonizing the VAE's latent structure with the classification objective.

\subsubsection{Knowledge Distillation}
\label{knowledge distillation}

As aforementioned, the GNN encoder may drift when trained on new tasks. To mitigate this, we incorporate knowledge distillation as a regularization term, applied to both current-task training sample $\V_{tr}^t$ and the exemplar set $\M^t$ from earlier tasks.

Let $\mathcal{D}_{\mathcal{M}}^t$ denotes the GNN embeddings for nodes in $\mathcal{M}^t$, the distillation loss is formulated as:
\begin{align}
    \LL_{\mathrm{kd}}^t =  \frac{1}{|\D^t_{\mathrm{real}}\cup\D^t_{\mathrm{\M}}|} \sum_{\mathbf{z}\in \mathcal{D}^t_{\mathrm{real}} \cup \mathcal{D}_{\mathcal{M}}^t} {\|\mathbf{z}_{\mathrm{teacher}} - \mathbf{z}_{\mathrm{student}} \|}^2 ,
\end{align}
where $\mathbf{z}_{\mathrm{teacher}}$ and $\mathbf{z}_{\mathrm{student}}$ are the GNN embedding from the previous task and current task, respectively. 

An alternative is to keep a pretrained GNN encoder fixed during incremental learning. However, unlike in image or text domains, graph pretraining typically occurs on the same dataset used downstream due to structural and feature diversity across domains (e.g., social vs. molecular graphs)~\cite{sun2023all,liu2023graphprompt,fang2024universal}. Thus, pretrained GNNs often struggle to generalize well when initial datasets are small. Instead, our method employs knowledge distillation, providing a flexible approach that preserves historical knowledge while slowly accommodating the evolving graph structures.


\subsubsection{Pseudo Sample Generation}
\label{pseudo id sample generation}

In this section, we describe how the ID samples are generated based on the prototypical CVAE. Given a prototype $p_c$ belonging to a class from the previous tasks, we sample latent variables $\mathbf{h}$ from the Gaussian prior $\mathcal{N}\bigl(\mathbf{p}_c, \mathbf{I}\bigr)$. These latent samples are then decoded by the CVAE decoder $p_{\theta}(\mathbf{z}|\mathbf{h})$ to reconstruct pseudo GNN embeddings $\hat{\mathbf{z}}$:
\begin{align}
    \mathbf{h} \sim \mathcal{N}\bigl(\mathbf{p}_c, \mathbf{I}\bigr), \quad \hat{\mathbf{z}} = p_{\theta}(\mathbf{z}|\mathbf{h}).
\end{align}
The generated embeddings $\hat{\mathbf{z}}$ are subsequently used as synthetic samples for the corresponding class $c$ in subsequent tasks.

\subsection{Out-of-distribution Sample Generation}
\label{ood sample generation}
In class-incremental open-set node classification, the model must accurately recognize nodes from classes unseen during training. Note that unknown samples may belong to several distinct classes and form their own clusters in the latent space. However, the uncertainty regarding the number of these classes and their structures complicates the establishment of robust decision boundaries. To address this, we introduce a mixing strategy to synthesize OOD samples that approximate unknown regions during training, enhancing the model’s ability to reject unseen classes effectively.


\subsubsection{Mixing Strategy}
\label{mixing strategy for ood}

To generate synthetic OOD samples, we combine embeddings from two distinct known classes, as mixed embeddings are unlikely to align with any single class prototype. Let $\mathbf{z}_1$ and $\mathbf{z}_2$ be two embeddings sampled from classes $c_1$ and $c_2$, or pseudo ID embedding generated according to Section~\ref{pseudo id sample generation} from previous classes, respectively. A synthetic embedding $\mathbf{z}_{\mathrm{mix}}$ is constructed as a linear combination of the two embeddings:
\begin{align}
    \mathbf{z}_{\mathrm{mix}} = \alpha * \mathbf{z}_1 + (1-\alpha) * \mathbf{z}_2.
\end{align}
Here, the mixing coefficient $\alpha$ is sampled from a Beta distribution $\alpha \sim \mathrm{Beta}(\beta, \beta)$, where $\beta$ controls the sharpness of the distribution. For $\beta=1$, $\alpha$ is uniformly distributed, while larger values of $\beta$ favor combinations closer to one of the two embeddings. 

\subsubsection{Prototypical HSC Loss}
\label{prototype HSC}

To differentiate pseudo unknown samples from known classes, a naive solution is to assign a single prototype for all unknown samples. However, this ignores the potential diversity among unknown classes, creating inadequate decision boundaries. This is especially problematic for synthetic OOD samples generated via mixing (Section~\ref{mixing strategy for ood}), as these samples blend features from multiple classes and cannot be meaningfully clustered. Inspired by~\cite{deepSVDD,hsc}, we instead propose a prototypical hypersphere classification loss that establishes flexible class boundaries. This allows in-class samples to cluster naturally, while effectively treating mixed and out-of-class embeddings as outliers, thereby naturally rejecting pseudo-OOD samples.


Given a prototype vector $\mathbf{p}_c \in \mathbb{R}^d$ representing the center of class $c$, 
the prototypical hypersphere classification loss is formulated as:
\begin{align}
    \mathcal{L}_{\mathrm{phsc}}^t
    &= \frac{1}{|\D^t \cup \D_{\M}^t|} \sum_{\mathbf{z} \in \D^t\cup \D_{\M}^t} \Bigl\{ \sum_{c\in C} \Bigl[
    (1 - y_c)\, \log\bigl(l\bigl(\mathbf{h}(\mathbf{z}), \mathbf{p}_c\bigr)\bigr) \notag \\
    &\quad - 
    y_c\, \log\Bigl(1 - l\bigl(\mathbf{h}(\mathbf{z}), \mathbf{p}_c\bigr)\Bigr)\Bigr] \Bigr\},
\end{align}
where $\mathbf{h}(\mathbf{z})$ denotes the CVAE encoded representation of $\mathbf{z}$, i.e., $\mathbf{h}(\mathbf{z}) = \mu_{\phi}(\mathbf{z})$. $C$ denotes the known class set, $y_c \in \{0,1\}$ indicates whether $\mathbf{h}$ belongs to class $c$ ($y_c=1$) or not ($y_c=0$), and $l(\mathbf{h}(\mathbf{z}), \mathbf{p}_c) = \exp(-\|\mathbf{h}(\mathbf{z}) - \mathbf{p}_c\|^2)$ represents a radial basis function, 
which measures the similarity between $\mathbf{h}(\mathbf{z})$ and $\mathbf{p}_c$. This function effectively defines a hypersphere centered at $\mathbf{p}_c$, where points closer to the prototype exhibit higher similarity scores.

For samples belonging to class $c$ ($y_c=1$), the loss encourages $\mathbf{h}(\mathbf{z})$ to align closely with the prototype $\mathbf{p}_c$. This promotes the formation of compact clusters around each class prototype in the embedding space. Conversely, for samples that do not belong to class $c$ ($y_c=0$), the loss effectively drives $\mathbf{h}(\mathbf{z})$ away from the hypersphere defined by $\mathbf{p}_c$. 
Therefore, $\mathcal{L}_{\mathrm{phsc}}$ naturally handles the pseudo OOD samples by treating them as negative ($y_c=0$) for all known-class prototypes. This ensures that OOD samples are repelled from every hypersphere, resulting in their placement in regions of the embedding space that are sufficiently distant from all known class prototypes. 

\subsection{Training Objective}
\label{overall_training}

Overall, the training loss for each task $\mathcal{T}^t$ can be written as:
\begin{align}
    \mathcal{L}^t = \mathcal{L}^t_{\mathrm{phsc}} + \mathcal{L}^t_{\mathrm{pcvae}} + \lambda_{\mathrm{kd}} \,\mathcal{L}^t_{\mathrm{kd}},
\end{align}
where $\lambda_{\mathrm{kd}}$ is the hyperparameter that balances the contribution of knowledge distillation. During the inference of each task $\mathcal{T}^t$, we compute an open-set score for each test node, which is defined as:
\begin{align}
    s_{\mathrm{open}}(\mathbf{z}) = -\min_{c\in C^t} \|\mathbf{z} - \mathbf{p}_c\|^2.
\end{align}
This score can be later thresholded to determine its association with known or unknown classes, i.e., 1 as known, and 0 as unknown.

\section{Experiments}
\subsection{Experimental Settings}

\subsubsection{Datasets}
\begin{table*}[t]
\centering
\caption{Performance comparison of each dataset over all tasks in terms of open-set classification rate (OSCR), closed-set accuracy (ACC), and open-set AUC-ROC (AUC). All results are averaged over 5 independent runs. }
\resizebox{\linewidth}{!}{
\begin{tabular}{c |  c c c | c c c | c c c | c c c | c c c }
\hline
 & \multicolumn{3}{c|}{\textbf{Photo}} & \multicolumn{3}{c|}{\textbf{Computer}}  & \multicolumn{3}{c|}{\textbf{CS}} & \multicolumn{3}{c}{\textbf{CoraFull}} & \multicolumn{3}{c}{\textbf{Arxiv}} \\ \hline
 
\textbf{Methods}  & \textbf{OSCR} & \textbf{ACC} & \textbf{AUC}& \textbf{OSCR} & \textbf{ACC} & \textbf{AUC}& \textbf{OSCR} & \textbf{ACC} & \textbf{AUC}& \textbf{OSCR} & \textbf{ACC} & \textbf{AUC}& \textbf{OSCR} & \textbf{ACC} & \textbf{AUC} \\
\hline

EWC (PNAS 2017)~\cite{ewc} & 0.640 & 0.862 & 0.703 & 0.533 & 0.804 & 0.620 & 0.639 & \textbf{0.811} & 0.736 & 0.430 & 0.541 & 0.698 & 0.338 & \textbf{0.461} & 0.541 \\

ODIN (ICLR 2018)~\cite{odin} & 0.450 & 0.851 & 0.514 & 0.442 & 0.802 & 0.544 & 0.437 & 0.796 & 0.539 & 0.291 & 0.524 & 0.534 & 0.230 & 0.426 & 0.447 \\

OpenWGL (ICDM 2020)~\cite{openWGL} & 0.687 & \textbf{0.871} & 0.753 & 0.567 & 0.845 & 0.656 & 0.540 & 0.782 & 0.647 & 0.424 & 0.559 & 0.672 & 0.349 & 0.383 & 0.574 \\

ERGNN (AAAI 2021)~\cite{ergnn} & 0.636 & 0.852 & 0.703 & 0.523 & 0.782 & 0.623 & 0.647 & 0.801 & 0.744 & 0.427 & 0.536 & 0.709 & 0.326 & 0.428 & 0.547 \\

IsoMax (TNNLS 2022)~\cite{old_isomax} & 0.565 & 0.778 & 0.692 & 0.521 & 0.830 & 0.612 & 0.516 & 0.744 & 0.639 & 0.373 & 0.504 & 0.605 & 0.342 & 0.363 & 0.578 \\

OpenWRF (IJCNN 2023)~\cite{openWRF} & 0.513 & 0.755 & 0.648 & 0.537 & 0.819 & 0.646 & 0.506 & 0.718 & 0.645 & 0.391 & 0.520 & 0.609 & 0.331 & 0.356 & 0.565 \\

SSRM (ICML 2023)~\cite{ssrm} & 0.597 & 0.831 & 0.678 & 0.530 & 0.791 & 0.629 & 0.614 & 0.791 & 0.722 & 0.458 & \textbf{0.577} & 0.702 & 0.327 & 0.431 & 0.542 \\

G$^2$Pxy (IJCAI 2023)~\cite{zhang2023g2pxy} & 0.584 & 0.762 & 0.719 & 0.529 & 0.695 & 0.675 & 0.627 & 0.693 & 0.759 & 0.343 & 0.450 & 0.645 & 0.340 & 0.349 & 0.597 \\

TPP (NeurIPS 2024)~\cite{tpp} & 0.413 & 0.569 & 0.631 & 0.400 & 0.594 & 0.585 & 0.623 & 0.722 & 0.754 & 0.326 & 0.435 & 0.605 & 0.370 & 0.364 & 0.643 \\

\hline

OGCIL (ours) & \textbf{0.750} & 0.869 & \textbf{0.835} & \textbf{0.667} & \textbf{0.868} & \textbf{0.766} & \textbf{0.658} & 0.808 & \textbf{0.762} & \textbf{0.484} & \textbf{0.577} & \textbf{0.720} & \textbf{0.409} & 0.414 & \textbf{0.651} \\

\hline
\end{tabular}}
\label{tab:comparison_result}
\end{table*}
\begin{figure*}[ht]
\begin{subfigure}[t]{1\linewidth}
    \centering
    \includegraphics[height=0.75cm]{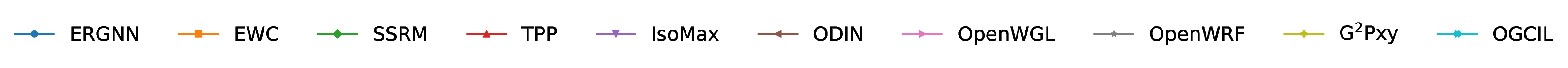}
\end{subfigure}
\vspace{-0.5cm}
\newline
\begin{subfigure}[b]{.19\linewidth}
    \centering
    \includegraphics[height=2.76cm]{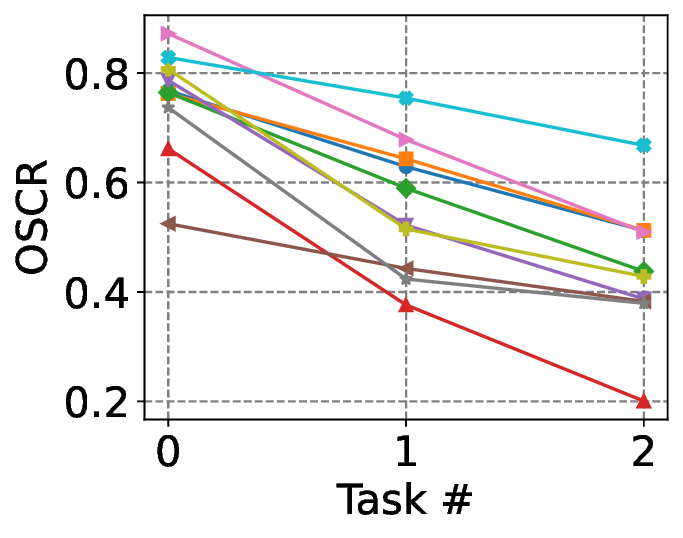} 
    \caption{Photo (OSCR)}
    \label{fig: oscr_photo}
\end{subfigure}
\hfill
\centering
\begin{subfigure}[b]{.19\linewidth}
    \centering
    \includegraphics[height=2.76cm]{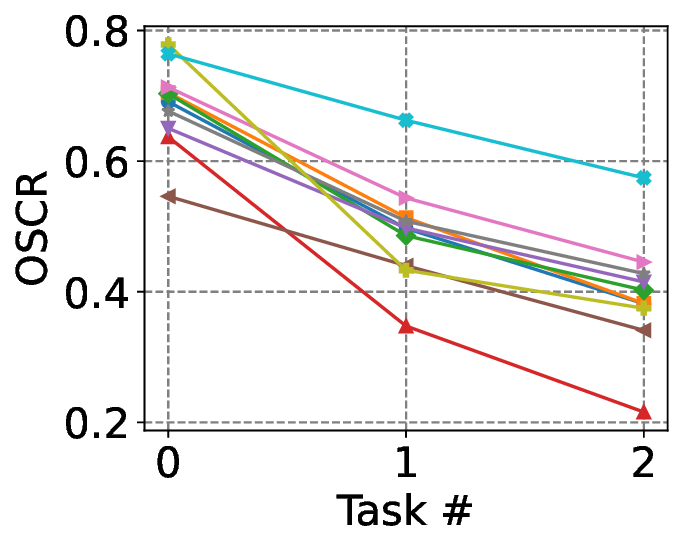}  
    \caption{Computer (OSCR)}
    \label{fig: oscr_computer}
\end{subfigure}
\hfill
\centering
\begin{subfigure}[b]{.19\linewidth}
    \centering
    \includegraphics[height=2.76cm]{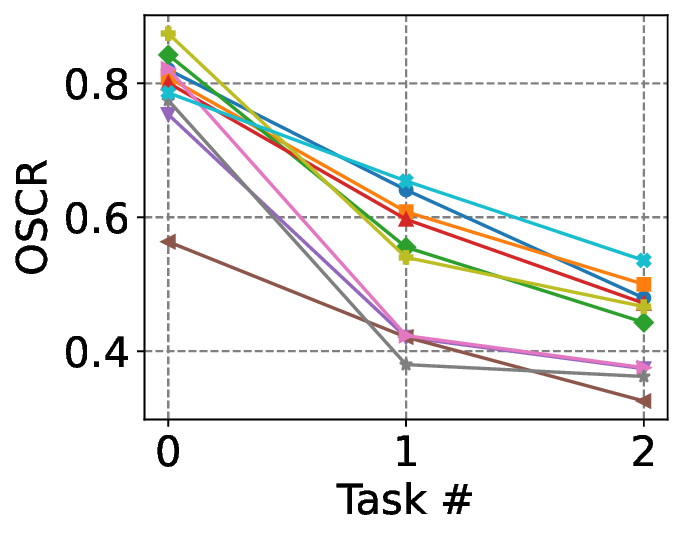}  
    \caption{CS (OSCR)}
    \label{fig: oscr_cs}
\end{subfigure}
\hfill
\centering
\begin{subfigure}[b]{.19\linewidth}
    \centering
    \includegraphics[height=2.76cm]{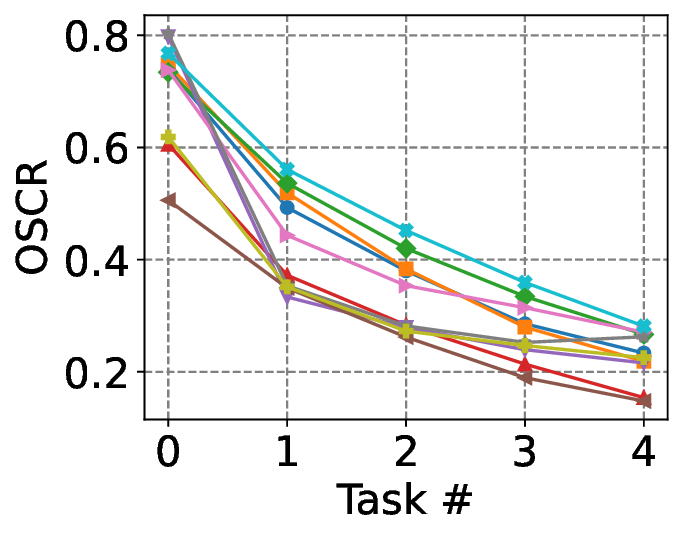}  
    \caption{CoraFull (OSCR)}
    \label{fig: oscr_corafull}
\end{subfigure}
\hfill
\centering
\begin{subfigure}[b]{.19\linewidth}
    \centering
    \includegraphics[height=2.76cm]{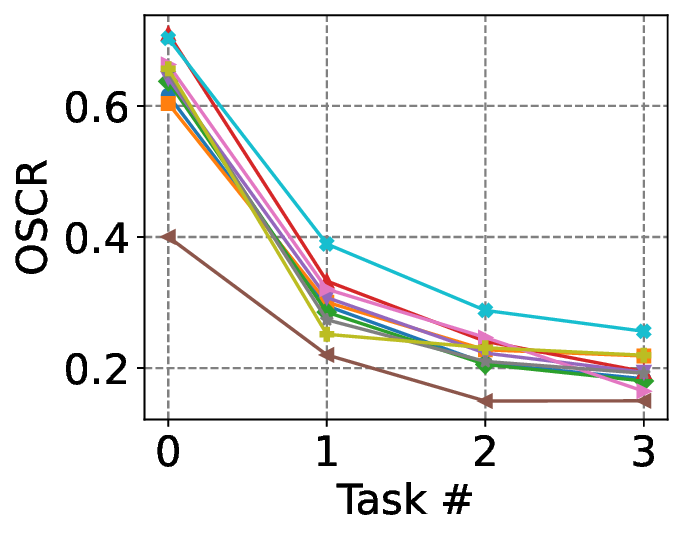}  
    \caption{Arxiv (OSCR)}
    \label{fig: oscr_arxiv}
\end{subfigure}
\newline
\begin{subfigure}[b]{.19\linewidth}
    \centering
    \includegraphics[height=2.76cm]{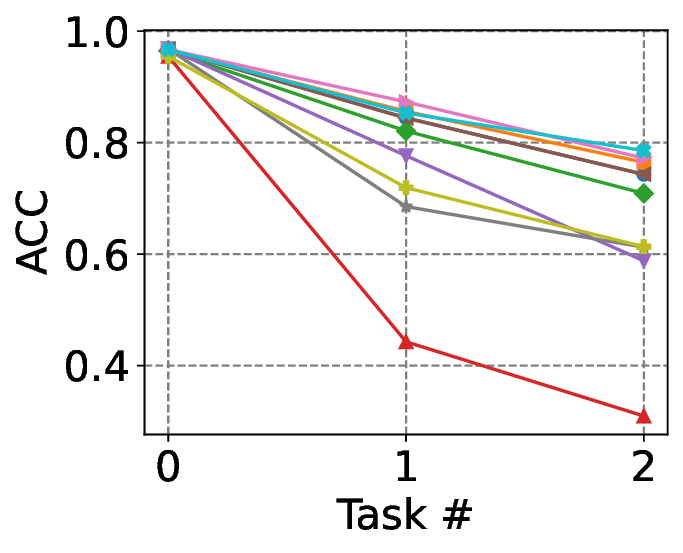} 
    \caption{Photo (ACC)}
    \label{fig: acc_photo}
\end{subfigure}
\hfill
\centering
\begin{subfigure}[b]{.19\linewidth}
    \centering
    \includegraphics[height=2.76cm]{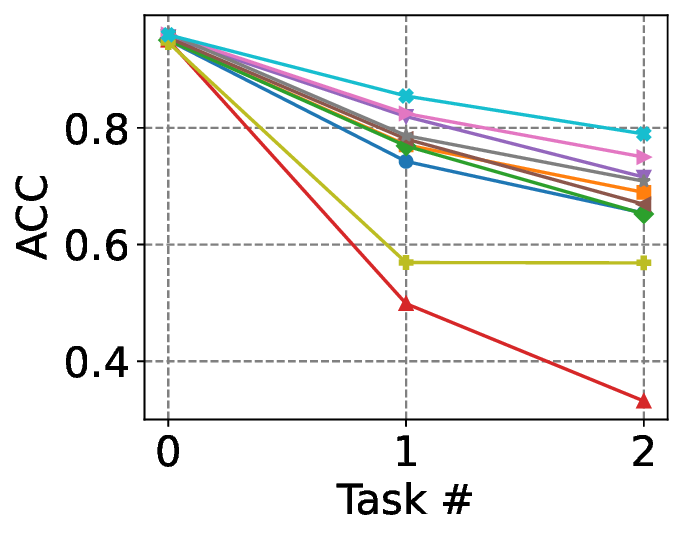}  
    \caption{Computer (ACC)}
    \label{fig: acc_computer}
\end{subfigure}
\hfill
\centering
\begin{subfigure}[b]{.19\linewidth}
    \centering
    \includegraphics[height=2.76cm]{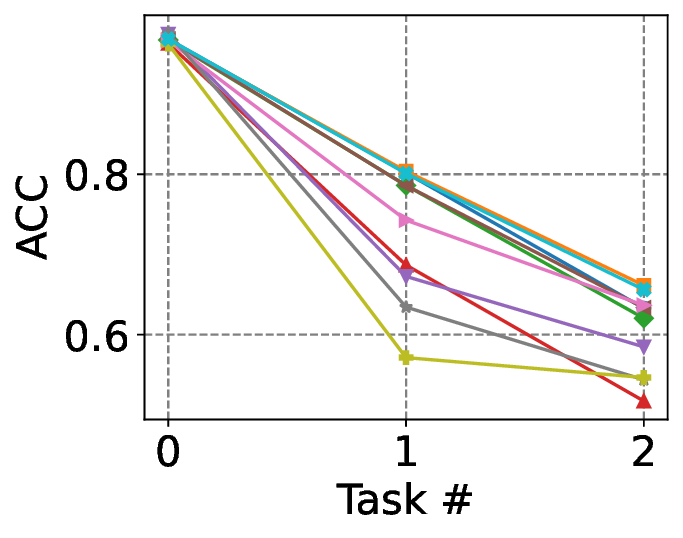}  
    \caption{CS (ACC)}
    \label{fig: acc_cs}
\end{subfigure}
\hfill
\centering
\begin{subfigure}[b]{.19\linewidth}
    \centering
    \includegraphics[height=2.76cm]{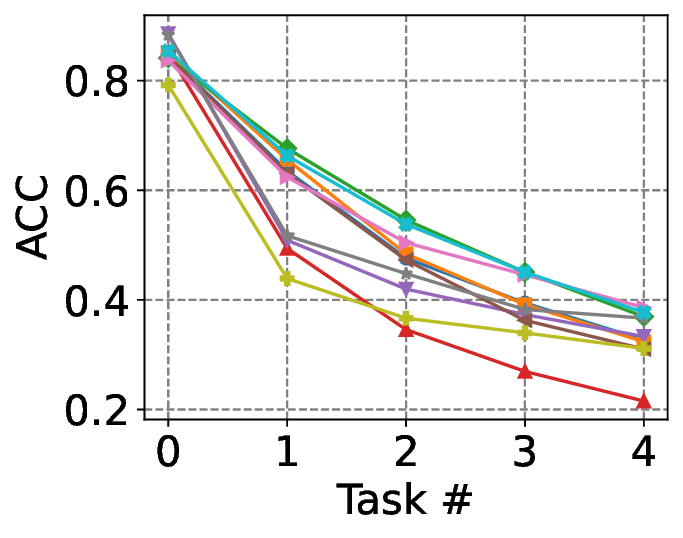}  
    \caption{CoraFull (ACC)}
    \label{fig: acc_corafull}
\end{subfigure}
\hfill
\centering
\begin{subfigure}[b]{.19\linewidth}
    \centering
    \includegraphics[height=2.76cm]{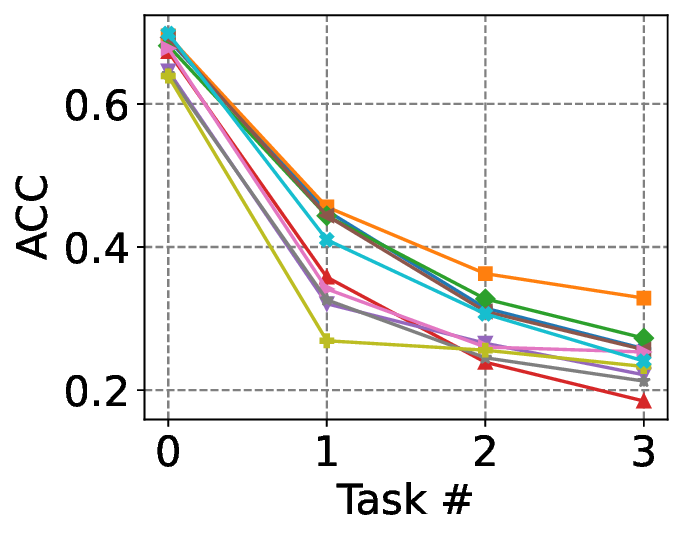}  
    \caption{Arxiv (ACC)}
    \label{fig: acc_arxiv}
\end{subfigure}
\newline
\begin{subfigure}[b]{.19\linewidth}
    \centering
    \includegraphics[height=2.76cm]{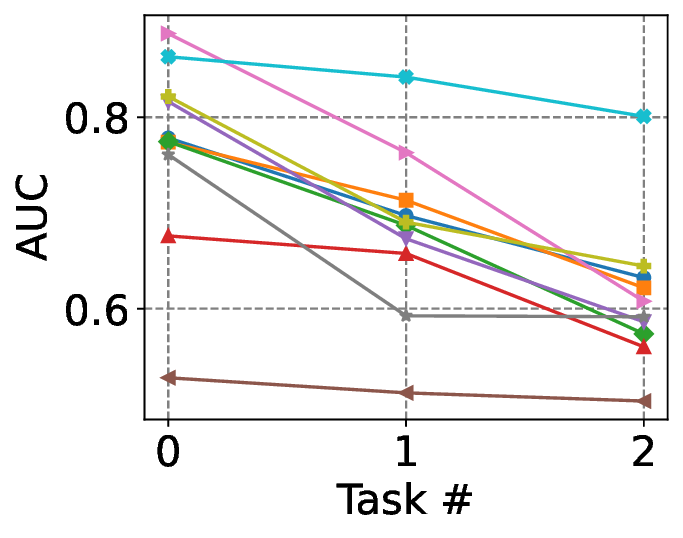} 
    \caption{Photo (AUC)}
    \label{fig: aucroc_photo}
\end{subfigure}
\hfill
\centering
\begin{subfigure}[b]{.19\linewidth}
    \centering
    \includegraphics[height=2.76cm]{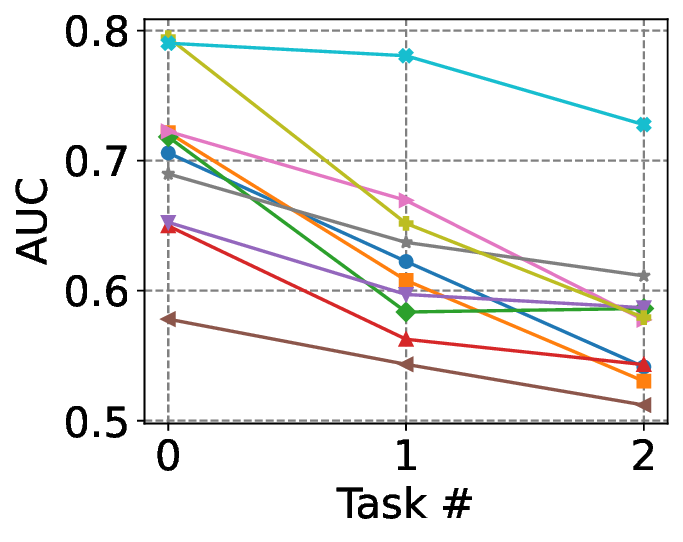}  
    \caption{Computer (AUC)}
    \label{fig: aucroc_computer}
\end{subfigure}
\hfill
\centering
\begin{subfigure}[b]{.19\linewidth}
    \centering
    \includegraphics[height=2.76cm]{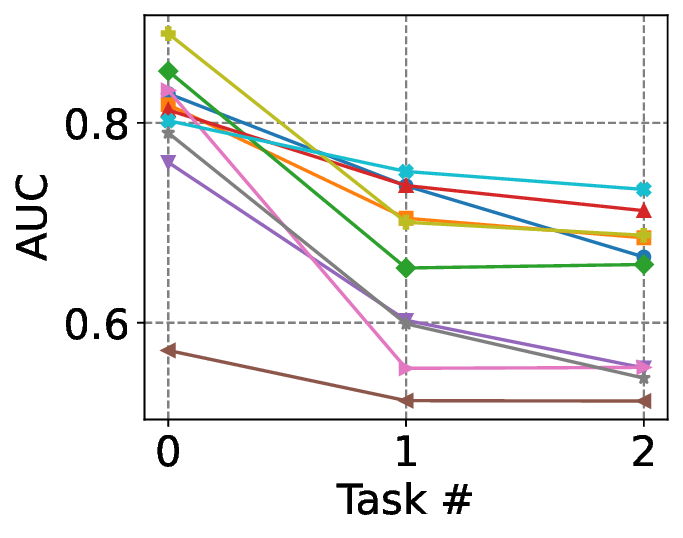}  
    \caption{CS (AUC)}
    \label{fig: aucroc_cs}
\end{subfigure}
\hfill
\centering
\begin{subfigure}[b]{.19\linewidth}
    \centering
    \includegraphics[height=2.76cm]{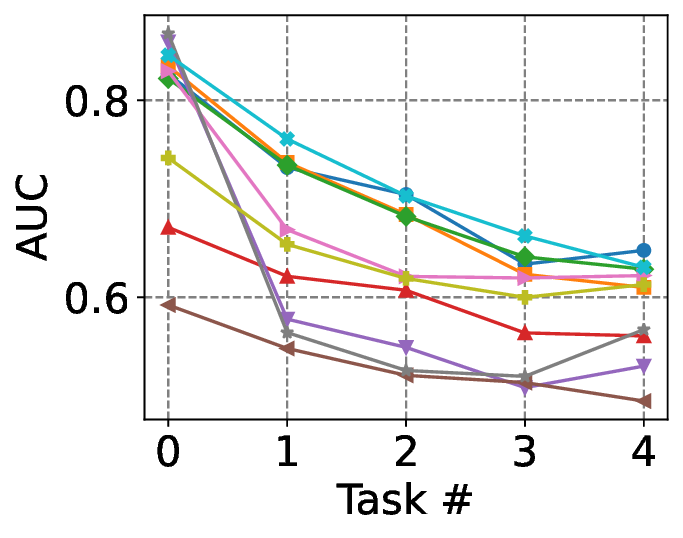}  
    \caption{CoraFull (AUC)}
    \label{fig: aucroc_corafull}
\end{subfigure}
\hfill
\centering
\begin{subfigure}[b]{.19\linewidth}
    \centering
    \includegraphics[height=2.76cm]{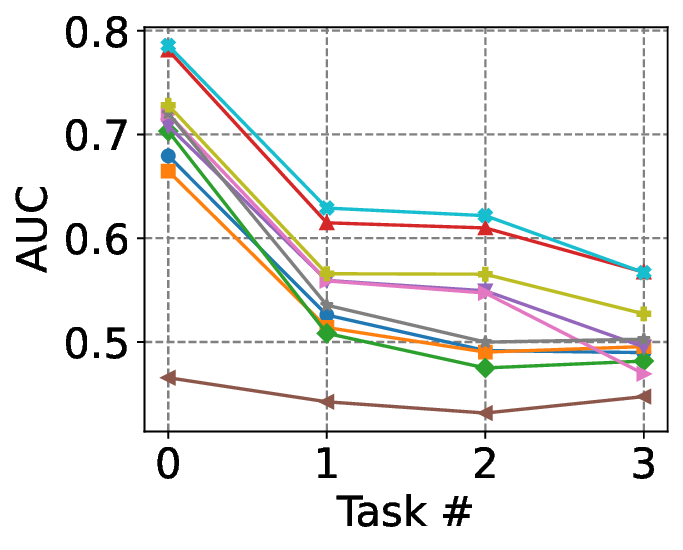}
    \caption{Arxiv (AUC)}
    \label{fig: aucroc_arxiv}
\end{subfigure}
\caption{Average performance comparison of each dataset over each task in terms of OSCR, ACC, and AUC.}
\label{fig:whole_comparison}
\end{figure*}

We evaluate our method on five datasets: CoraFull~\cite{corafull}, Computer~\cite{computerphoto}, Photo~\cite{computerphoto}, CS~\cite{computerphoto}, and Arxiv~\cite{arxiv}. Each dataset is divided into sequential tasks with disjoint classes, divided into known and unknown categories. For data efficiency, we allow unknown classes in one task to become known in subsequent tasks. 
This division enables the creation of more tasks but also mimics real-world scenarios where initially unseen classes emerge as known.
Under our inductive setting, nodes from unknown classes are masked during training, ensuring no prior exposure. For data splits, we adopt the original test partition for Arxiv, while using a uniform 40\%/20\%/40\% split (training/validation/test) for all other datasets that do not provide a test partition.


\begin{figure*}
\includegraphics[height=3.68cm]{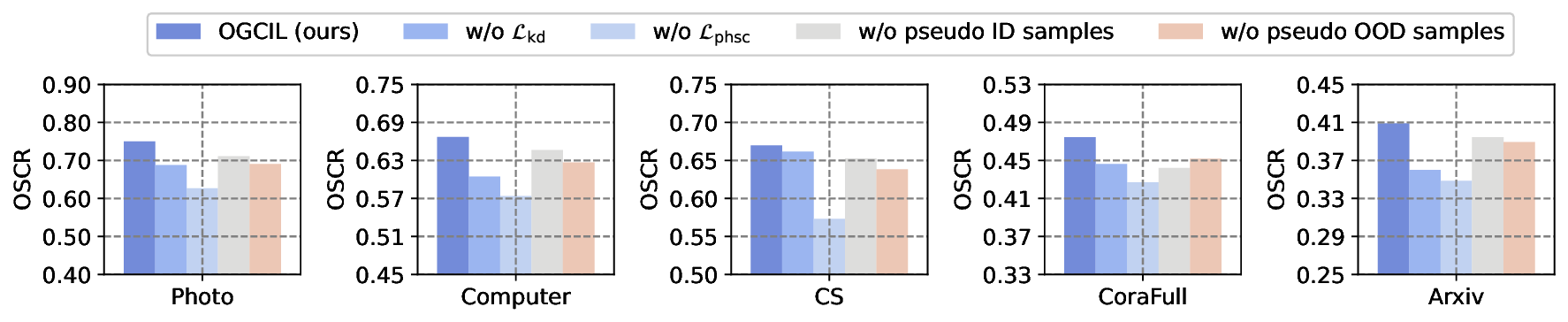}
\caption{Ablation study in terms of average open-set classification rate (OSCR) over all tasks.}
\hfill
\label{fig:ablation}
\end{figure*}

\subsubsection{Evaluation Metric}

We evaluate methods using three metrics: Open-Set Classification Rate (OSCR)~\cite{oscr}, Closed-Set Accuracy, and Open-Set AUC-ROC. 
Closed-Set Accuracy assesses performance purely on known classes. Open-Set AUC-ROC evaluates how well models distinguish between known and unknown classes without relying on a specific threshold. OSCR is a threshold agnostic metric that measures the trade-off between correctly classifying known samples and incorrectly accepting unknown samples. 

\subsubsection{Baseline Methods and Implementation Details}

Given the absence of direct baselines tailored to our problem, we examine two categories of approaches: class-incremental approaches (EWC~\cite{ewc}, ERGNN~\cite{ergnn}, SSRM~\cite{ssrm}, TPP~\cite{tpp}) and open-set recognition methods (ODIN~\cite{odin}, IsoMax~\cite{old_isomax}, OpenWGL~\cite{openWGL}, G$^2$Pxy~\cite{zhang2023g2pxy}, OpenWRF~\cite{openWRF}). To adapt the class-incremental methods for open-set scenarios, we employ a widely accepted softmax-based score~\cite{zhang2023g2pxy,openWGL}, which designates samples as unknown if all class probabilities fall below a threshold. To address the absence of catastrophic forgetting mechanisms in open-set methods, we adapt all baselines with a replay strategy that retains exemplars consisting of a small subset of old-class samples. 
For exemplar selection, we primarily use the coverage maximization (CM) strategy of ERGNN~\cite{ergnn}, which selects nodes that maximize embedding-space coverage. For consistency, we retain 5 exemplars per class in the main experiments, and we adopt GCN~\cite{kipf2017semi} as the standard backbone for all methods for computational efficiency. Additionally, we experiment with the mean-of-features (MF) selection approach~\cite{ergnn}, the varying number of exemplars, and GAT as the backbone as discussed in Section~\ref{sec:hyperparam}.


\subsection{Comparison Study}

Table~\ref{tab:comparison_result} shows the average performance across five datasets, and Figure~\ref{fig:whole_comparison} illustrates detailed task-wise comparisons. OGCIL consistently achieves superior OSCR, outperforming the best competitor by up to 17.6\%, and demonstrates robust known-unknown separation through higher open-set AUC-ROC scores. Although our closed-set accuracy isn't always highest, it remains competitive, indicating no significant compromise from our open-set design.


Compared to class-incremental baselines, OGCIL significantly excels in OSCR due to dedicated open-set detection mechanisms. Class-incremental baselines fall short in detecting unknown samples because they lack explicit unknown detection, leading to overconfident known-class predictions. Conversely, OGCIL employs prototypical HSC loss combined with pseudo-unknown sample generation, creating tighter and more accurate decision boundaries. Our prototypical CVAE further enriches older-class representations through pseudo-ID embedding replay, refining classifier boundaries and enhancing both open-set and closed-set performance.


In comparison with other open-set baselines with replay mechanisms added, OGCIL uniquely generates pseudo-ID embeddings in latent space, effectively preserving historical knowledge without additional storage overhead. Additionally, Our HSC loss avoids forcing unknowns into rigid clusters, maintaining stringent boundaries for known classes while letting unknowns remain outside those boundaries. On the other hand, OpenWGL maximizes model uncertainty for low-confidence samples, which can inadvertently classify known samples as unknown. The problem exegerates particularly under the inductive setting, where unknown classes do not appear during training. Similarly, G$^2$Pxy consolidates all pseudo OOD samples into a single group, overlooking the inherent diversity of unseen classes and the potential noise in pseudo samples.


\subsection{Ablation Study}

We systematically ablate key components of OGCIL (Figure~\ref{fig:ablation}) to understand their individual contributions.
Removing knowledge distillation ($\mathcal{L}_{kd}$) leads to severe representation drift across tasks, diminishing the effectiveness of pseudo ID samples and, thus degrading the performance notably.
\begin{figure*}[t]
\vspace{-0.5cm}
\begin{subfigure}[t]{1\linewidth}
    \centering
    \includegraphics[height=0.8cm]{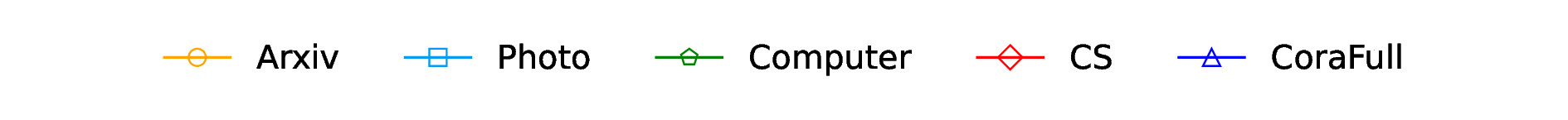}
\end{subfigure}
\vspace{-0.6cm}
\newline
\begin{subfigure}[b]{.195\linewidth}
    \includegraphics[height=2.73cm]{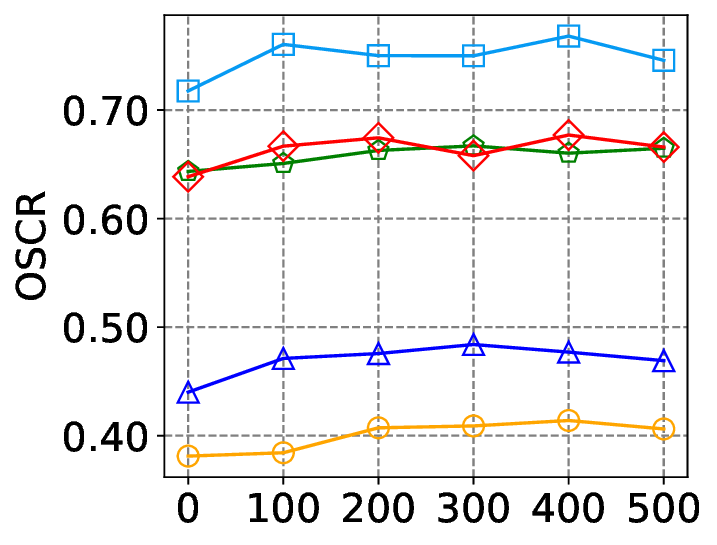} 
    \centering
    \caption{\# ID samples}
    \label{fig:id}
\end{subfigure}
\begin{subfigure}[b]{.195\linewidth}
    \includegraphics[height=2.73cm]{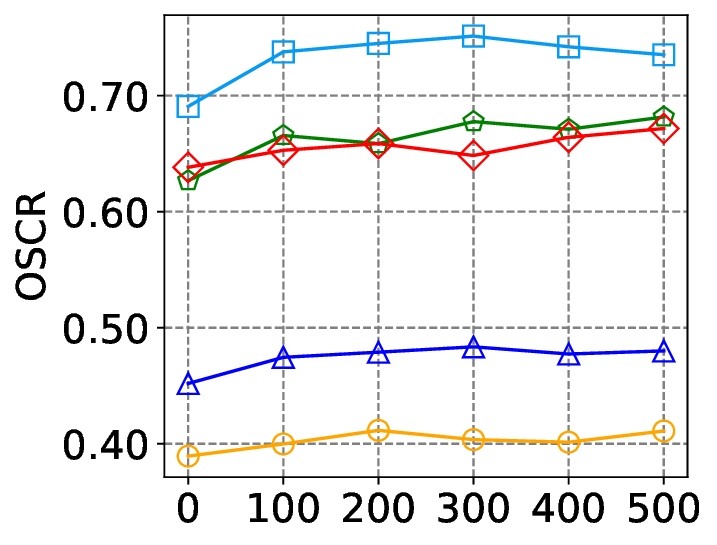}  
    \centering
    \caption{\# OOD samples}
    \label{fig:ood}
\end{subfigure}
\begin{subfigure}[b]{.195\linewidth}
    \includegraphics[height=2.73cm]{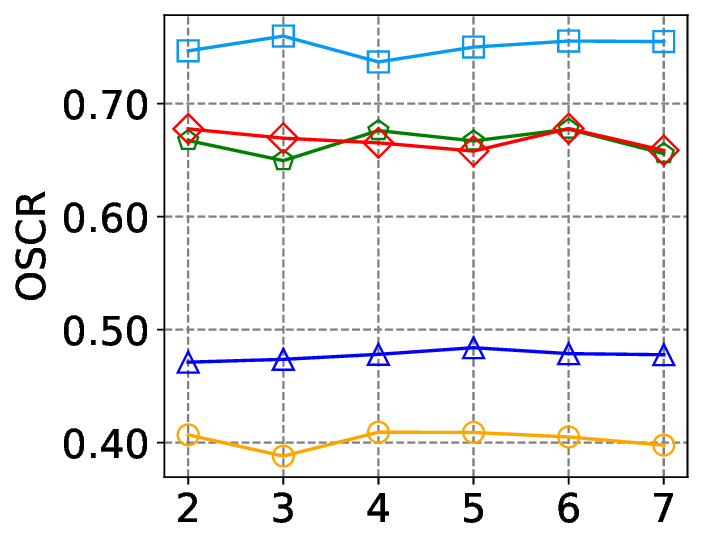}  
    \caption{$\beta$ in Beta distribution}
    \label{fig:beta}
\end{subfigure}
\begin{subfigure}[b]{.195\linewidth}
    \includegraphics[height=2.73cm]{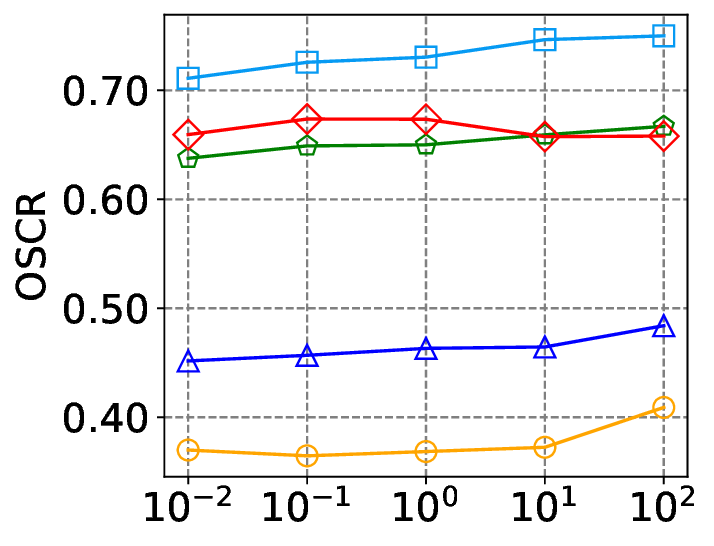}  
    \centering
    \caption{$\mathcal{L}_{\mathrm{kd}}$ weight $\lambda_{\mathrm{kd}}$}
    \label{fig:kd}
\end{subfigure}
\begin{subfigure}[b]{.195\linewidth}
    \includegraphics[height=2.73cm]{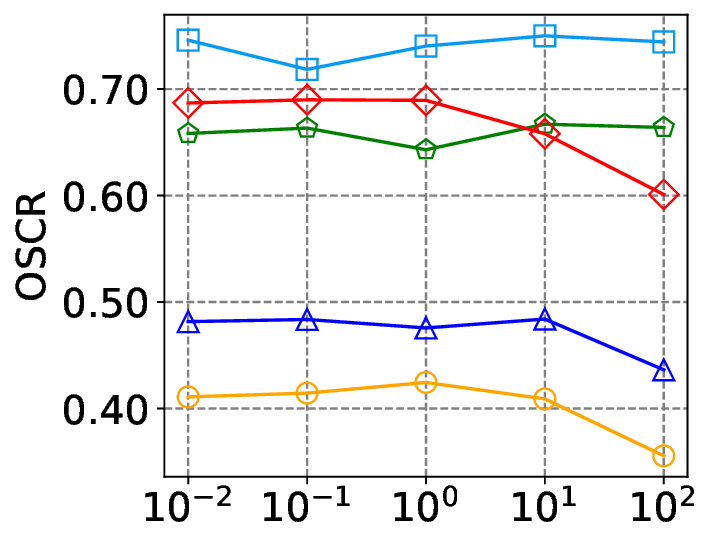} 
    \centering
    \caption{CVAE weight $\lambda_{\mathrm{reconst}}$}
    \label{fig:reconst}
\end{subfigure}
\newline
\begin{subfigure}[t]{1\linewidth}
    \centering
    \includegraphics[height=0.85cm]{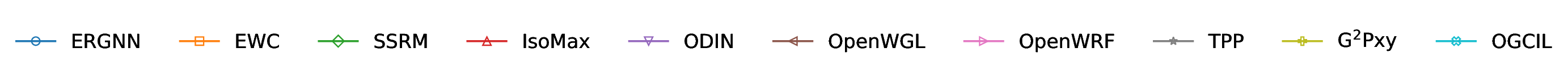}
\end{subfigure}
\vspace{-0.6cm}
\newline
\begin{subfigure}[b]{.195\linewidth}
    \includegraphics[height=2.7cm]{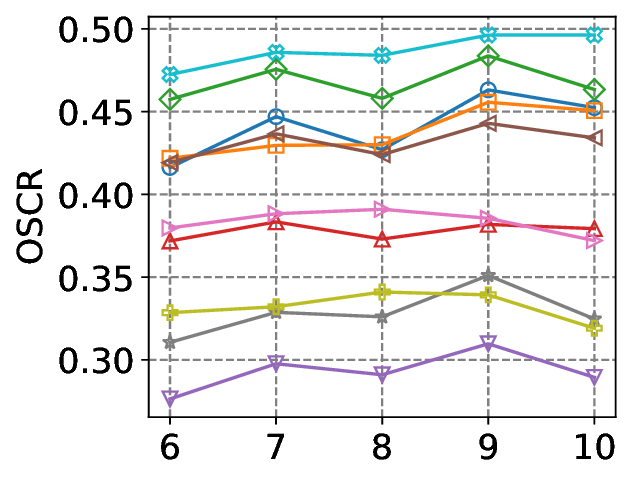} 
    \centering
    \caption{\# known classes/task}
    \label{fig:known_classes}
\end{subfigure}
\begin{subfigure}[b]{.195\linewidth}
    \includegraphics[height=2.7cm]{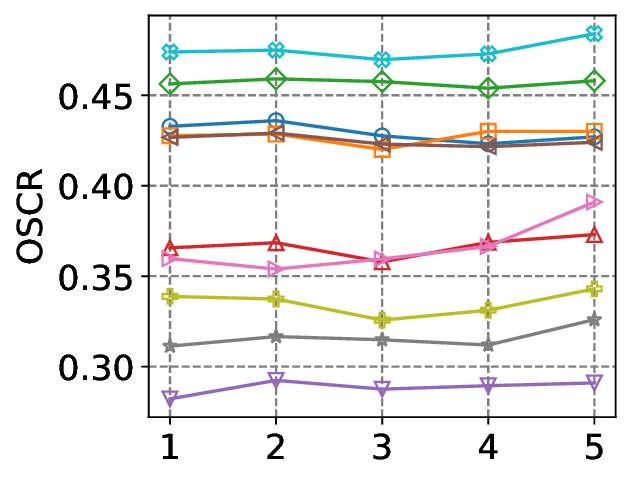} 
    \centering
    \caption{\# unknown classes/task}
    \label{fig:unknown_class}
\end{subfigure}
\begin{subfigure}[b]{.195\linewidth}
    \includegraphics[height=2.7cm]{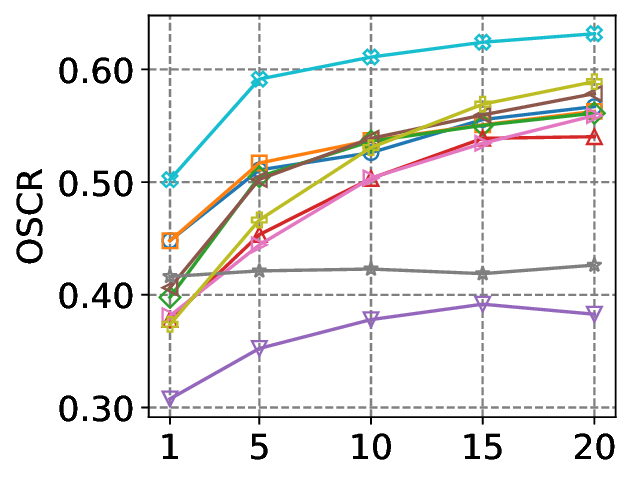}  
    \centering
    \caption{\# exemplars/task}
    \label{fig:num_examplar}
\end{subfigure}
\begin{subfigure}[b]{.195\linewidth}
    \includegraphics[height=2.7cm]{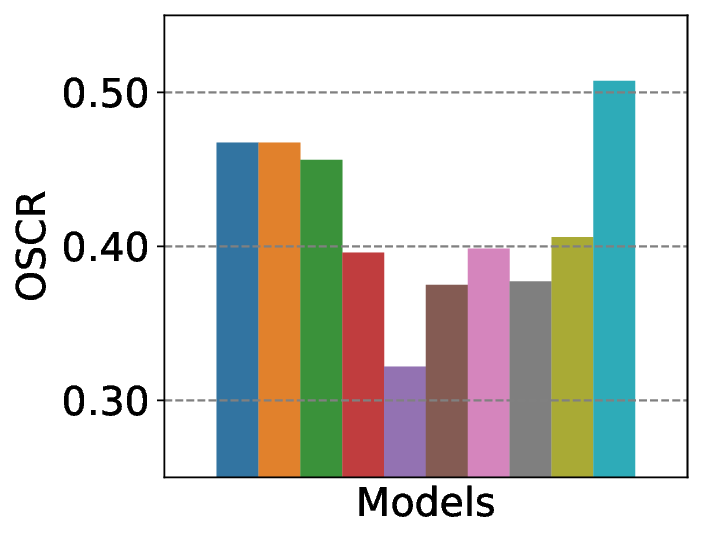}  
    \caption{MF as exemplar method}
    \label{fig:mf}
\end{subfigure}
\begin{subfigure}[b]{.195\linewidth}
    \includegraphics[height=2.7cm]{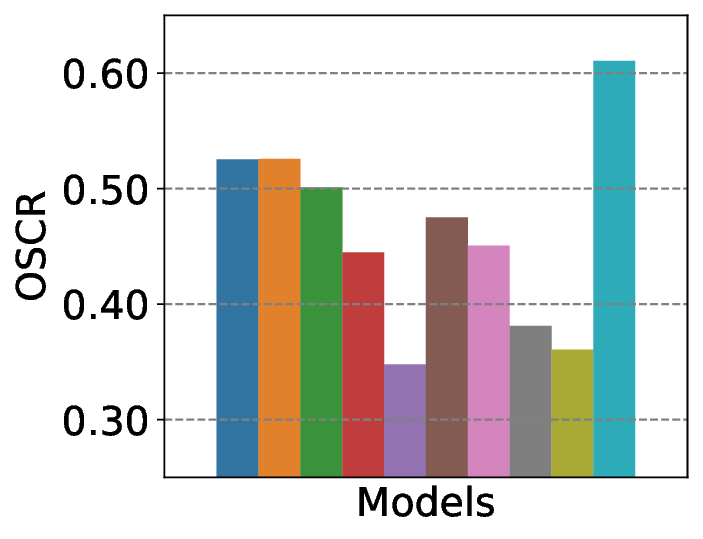}  
    \centering
    \caption{GAT as GNNs encoder}
    \label{fig:gat}
\end{subfigure}
\caption{Performance comparison in terms of average open-set classification rate (OSCR) versus different configurations.}
\label{fig:hyper_param}
\end{figure*}
\begin{figure*}[ht]
\hfill
\begin{subfigure}[b]{.16\linewidth}
    \includegraphics[height=2.2cm]{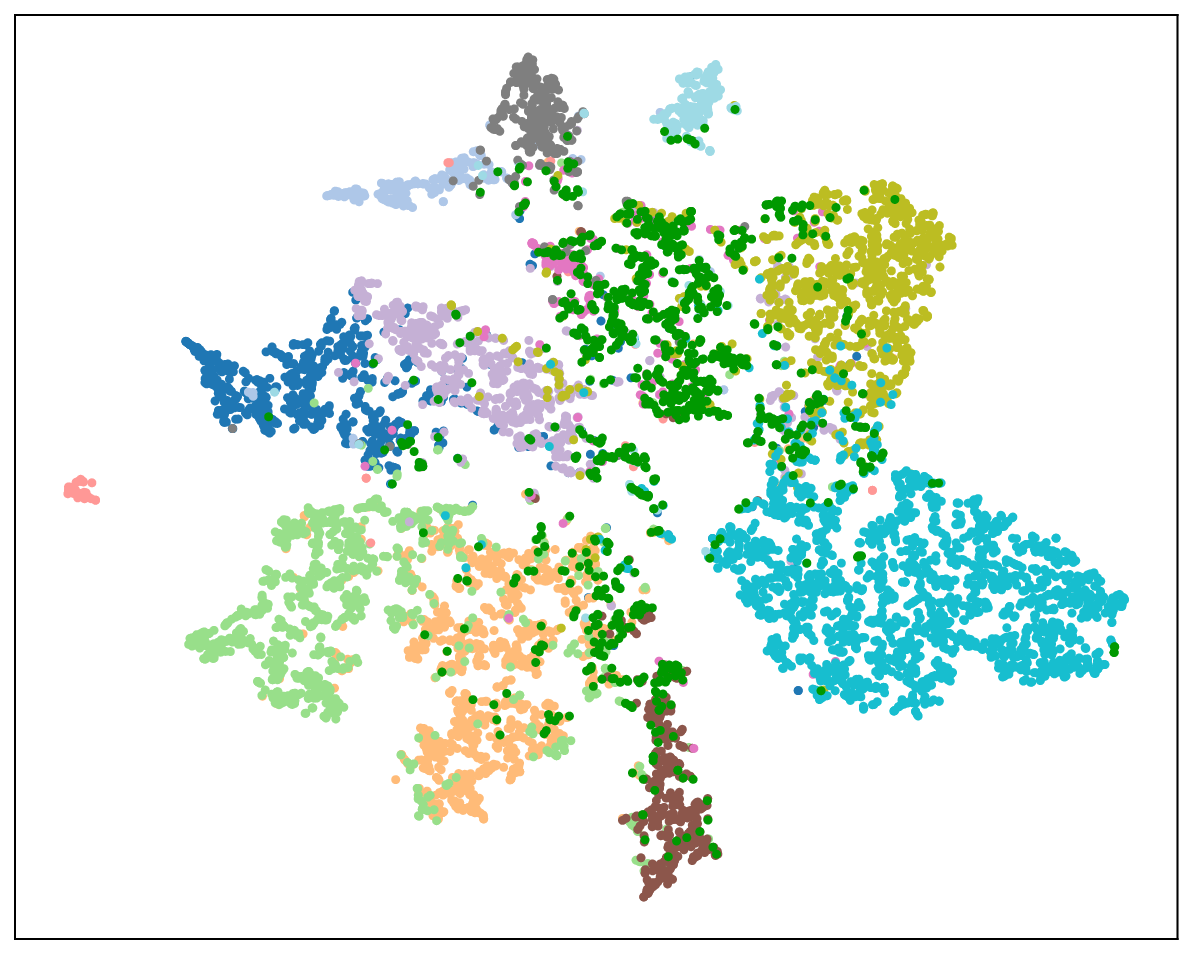} 
    \centering
    \caption{OGCIL (ours)}
    \label{fig:mymodel_visualization}
\end{subfigure}
\begin{subfigure}[b]{.16\linewidth}
    \includegraphics[height=2.2cm]{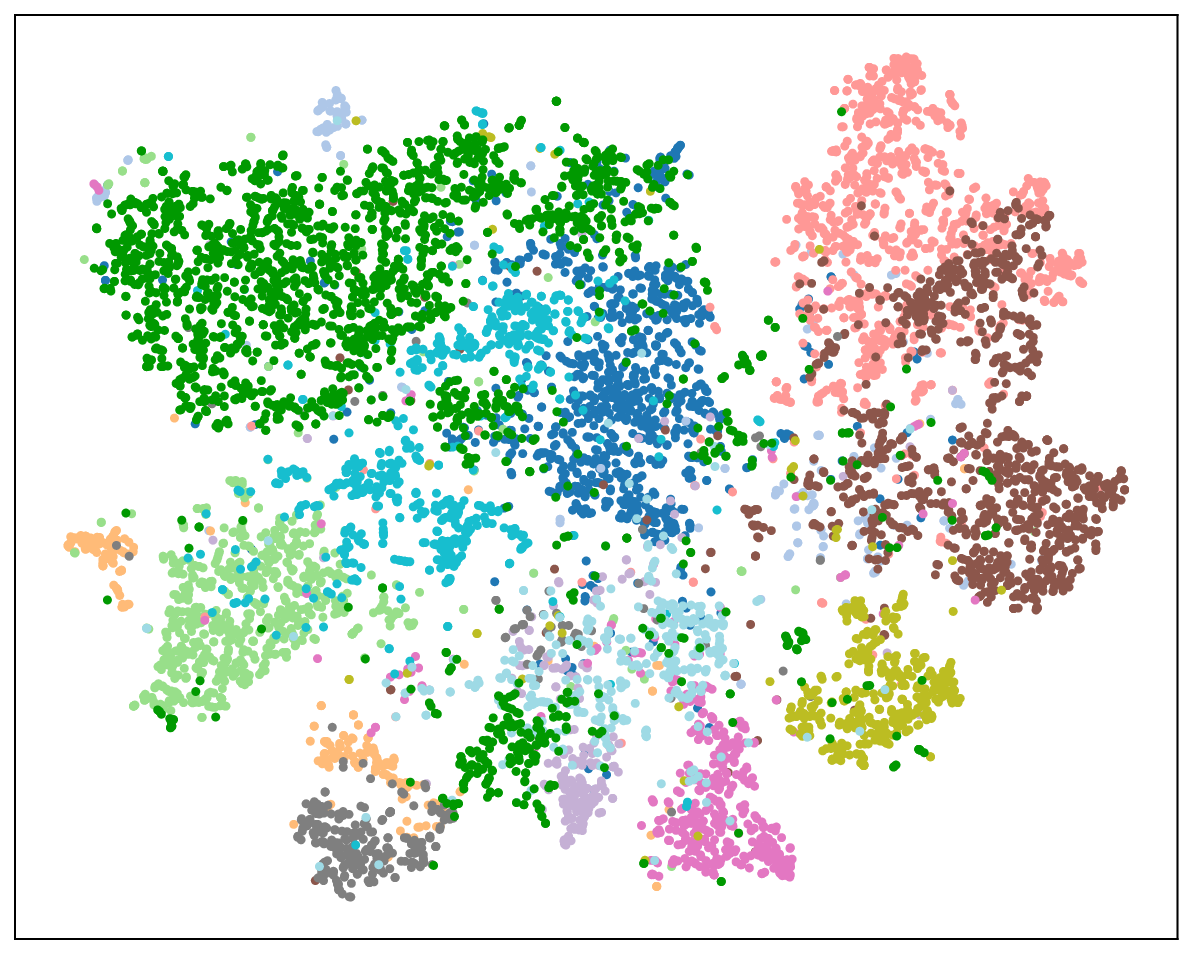} 
    \centering
    \caption{OpenWGL}
    \label{fig:openWGL_visualization}
\end{subfigure}
\begin{subfigure}[b]{.16\linewidth}
    \includegraphics[height=2.2cm]{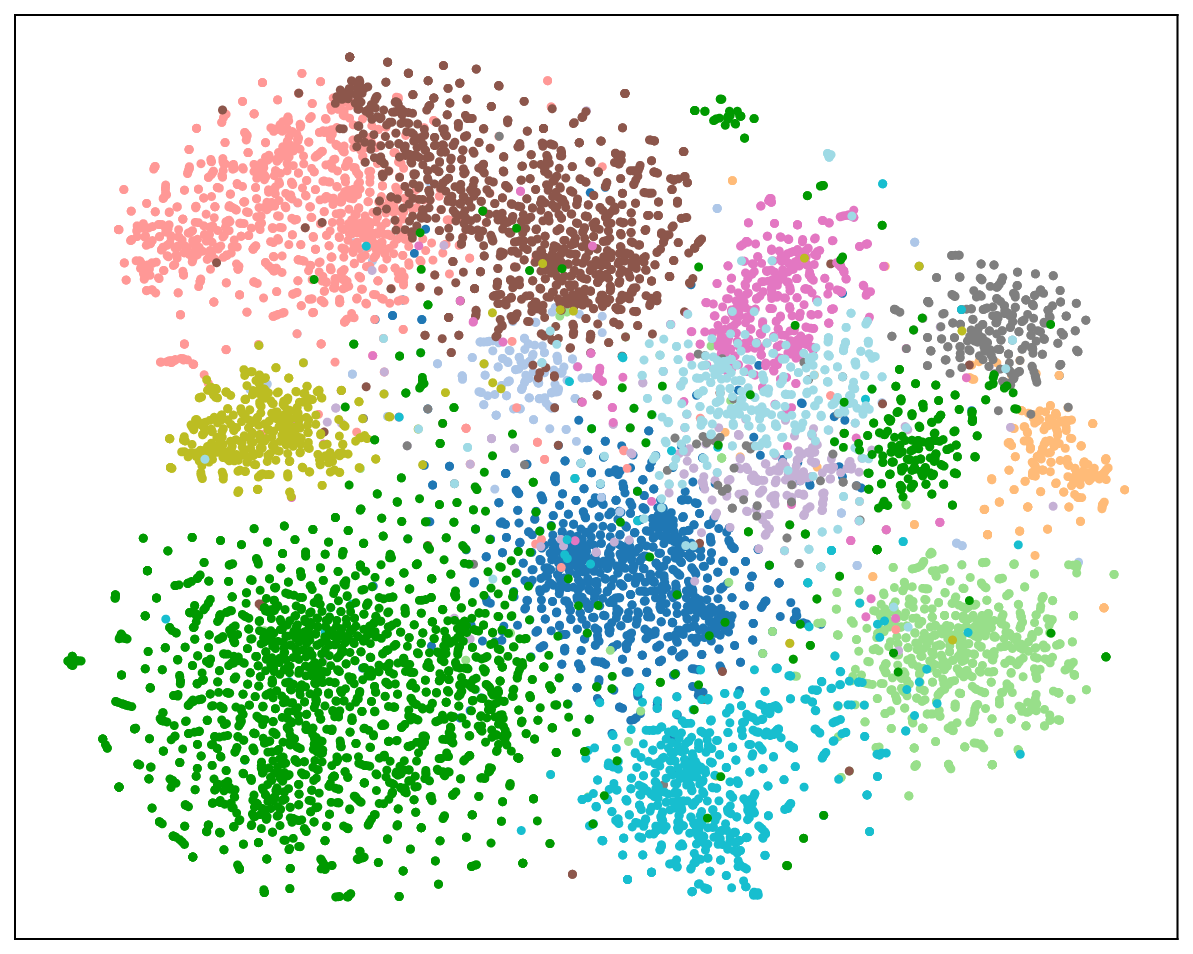} 
    \centering
    \caption{G$^2$Pxy}
    \label{fig:g2pxy_visualization}
\end{subfigure}
\begin{subfigure}[b]{.16\linewidth}
    \includegraphics[height=2.2cm]{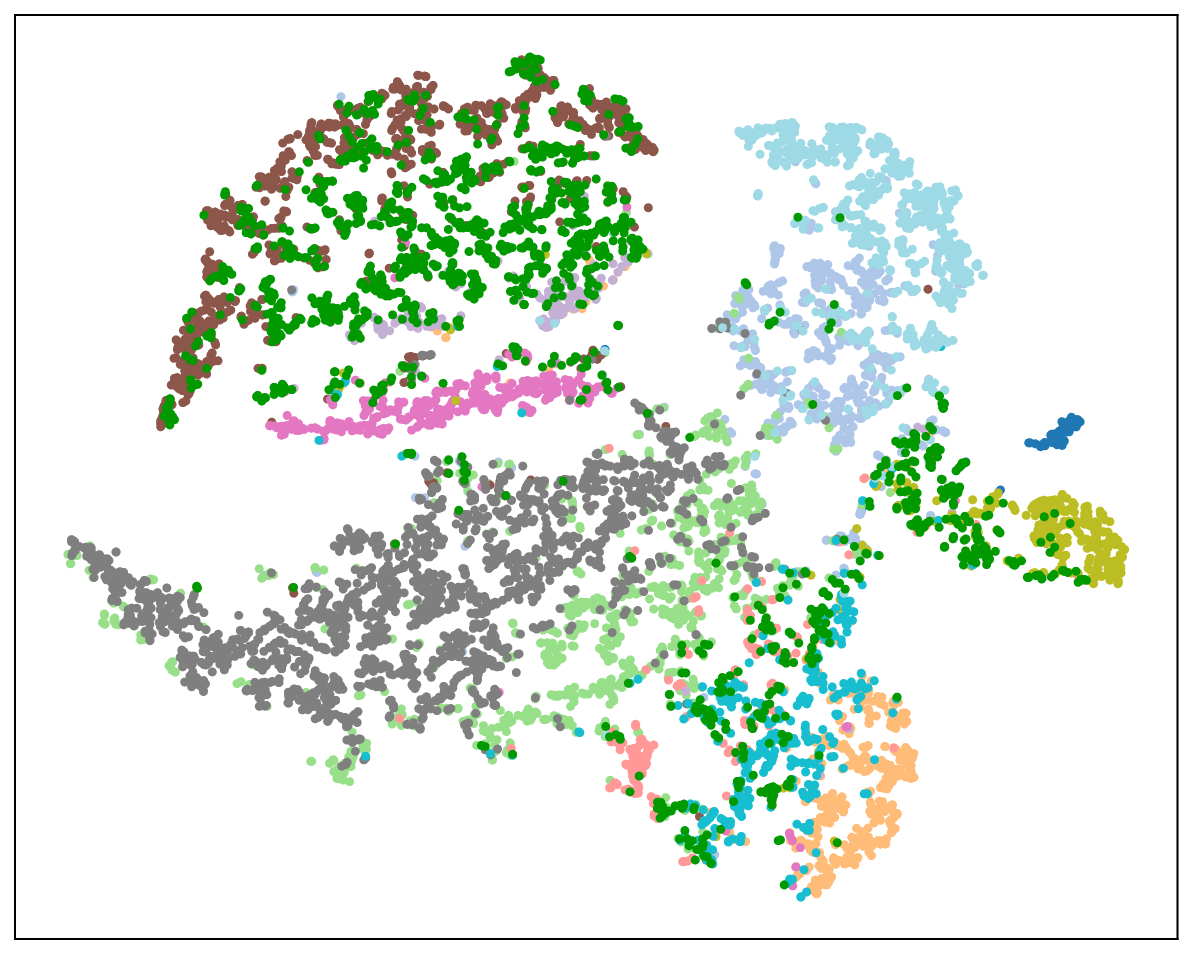} 
    \centering
    \caption{ERGNN}
    \label{fig:ergnn_visualization}
\end{subfigure}
\begin{subfigure}[b]{.16\linewidth}
    \includegraphics[height=2.2cm]{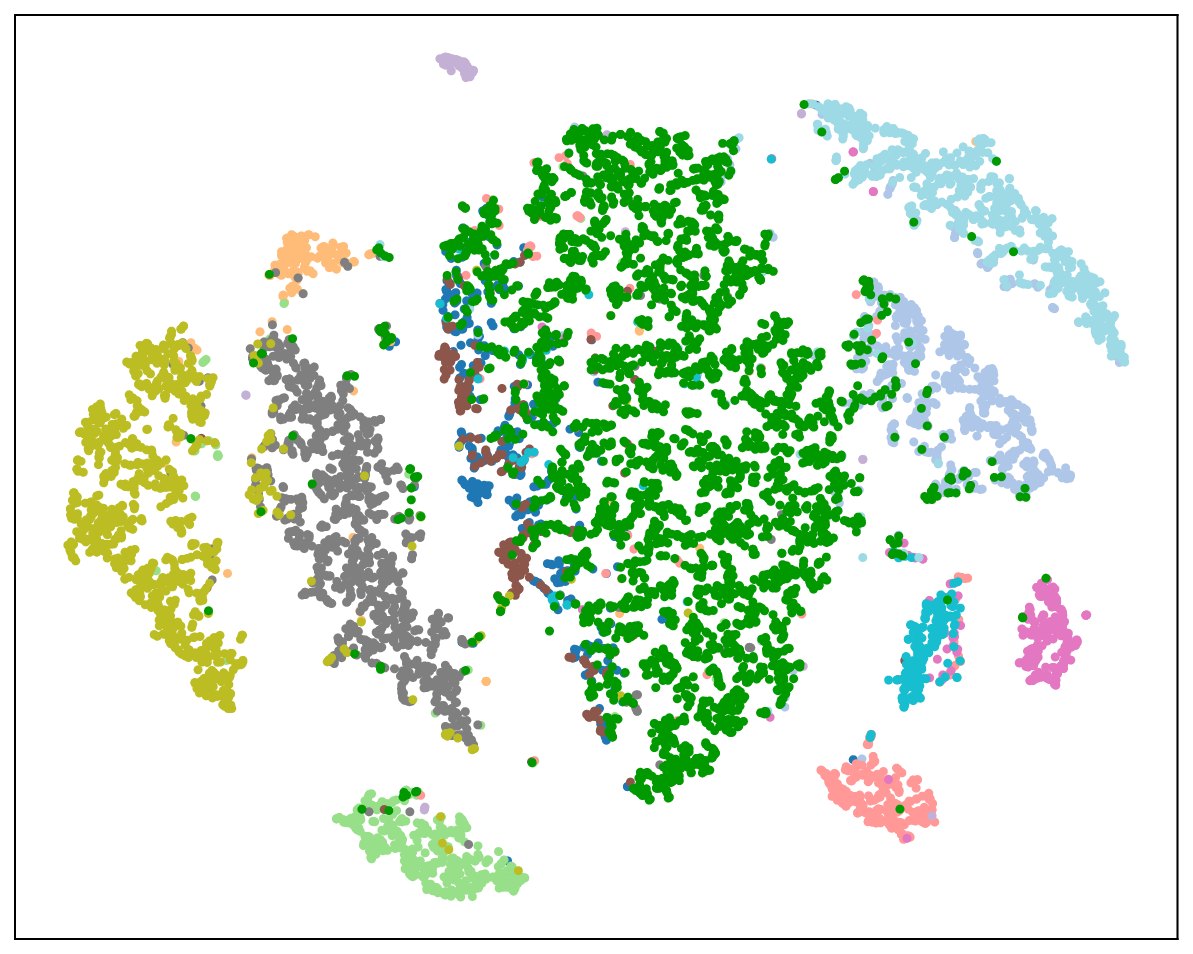} 
    \centering
    \caption{EWC}
    \label{fig:ewc_visualization}
\end{subfigure}
\begin{subfigure}[b]{.16\linewidth}
    \includegraphics[height=2.2cm]{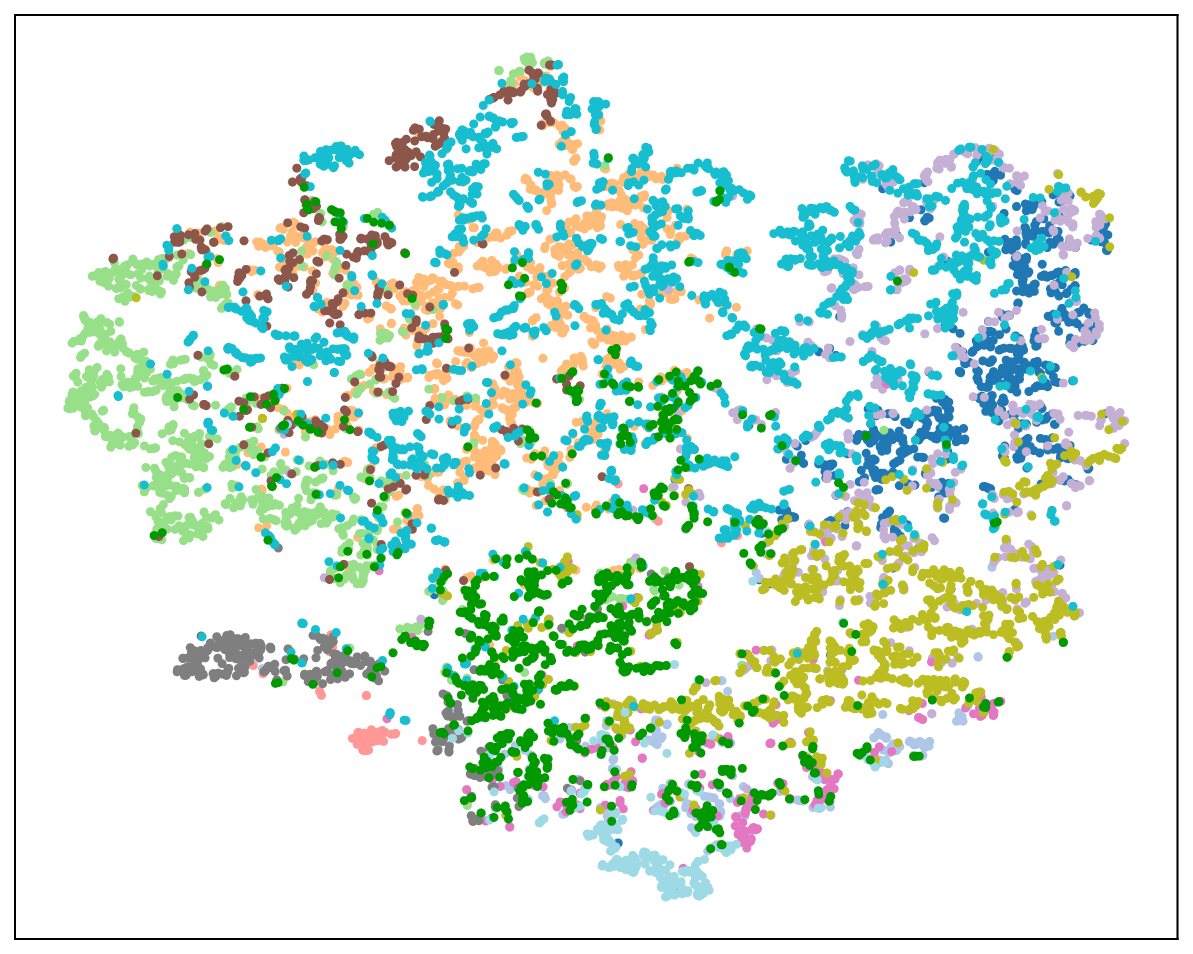} 
    \centering
    \caption{SSRM}
    \label{fig:ssrm_visualization}
\end{subfigure}
\caption{t-SNE embedding of CS dataset over the last task. Dark green denotes unknown classes.}
\label{fig:tsne}
\end{figure*}
Next, we replace the prototypical HSC loss ($\mathcal{L}_{phsc}$) with a standard distance-based cross-entropy classifier that assigns all OOD samples to a single “unknown” prototype. This replacement severely undermines the performance, as it fails to account for the diversity of unknown classes and the noisy, non-clustered nature of mixed OOD samples. Moreover, excluding pseudo ID samples forces reliance on a limited exemplar set, resulting in catastrophic forgetting of previously learned classes. 
Lastly, excluding pseudo OOD samples prevents the classifier from encountering samples from unknown classes, causing misclassification due to model overconfidence on known samples. These results confirm that each component is essential to balancing open-set recognition and incremental learning.

\subsection{Impact of Hyperparameters and Design}
\label{sec:hyperparam}

We conduct a comprehensive hyperparameter analysis to evaluate the impact of various factors, as shown in Figure~\ref{fig:hyper_param}. 

\subsubsection{Pseudo Sample Configurations}

Figures~\ref{fig:id} and~\ref{fig:ood} show that the number of ID and OOD samples do have impacts on the performance. Insufficient samples provide inadequate information about past tasks or unknown classes, thus degrading performance. Whereas excessively large numbers introduce noise, leading to a slight decline in OSCR. A balanced sample count ensures optimal results. Additionally, Figure~\ref{fig:beta} demonstrates that $\beta$, which controls the Beta distribution, has a stable impact on OSCR, suggesting that our model is robust to variations in this parameter.

\subsubsection{Loss Weight Analysis}

Figure~\ref{fig:kd} demonstrates the effect of $\lambda_{\mathrm{kd}}$. The increase of $\lambda_{\mathrm{kd}}$ improves OSCR, as higher values help stabilize the GNNs across tasks. This makes the generated pseudo embeddings more suitable for subsequent GNNs and prevents catastrophic forgetting. Moreover, Figure~\ref{fig:reconst} analyzes the impact of $\lambda_{\mathrm{reconst}}$. Specifically, excessively high values cause the model to overemphasize reconstruction at the expense of classification. Conversely, low values degrade the quality of pseudo embeddings. Thus, a moderate value of $\lambda_{\mathrm{reconst}}$ ensures a balance between reconstruction quality and classification performance.

\subsubsection{Effects from Different Class Composition}

Figures~\ref{fig:known_classes} and~\ref{fig:unknown_class} examine the effects of varying the number of known and unknown classes per task on the CoraFull dataset. Less known classes per task increase the total number of tasks, while increasing unknown classes introduces more diverse open-set conditions. Despite these variations, OGCIL consistently maintains robust performance, highlighting its adaptability to different task configurations.

\subsubsection{Effects from the Number of Examplars}

Figure~\ref{fig:num_examplar} evaluates the impact of exemplar count per class with performance averaged over five datasets. The performance of OGCIL rises quickly as the number of exemplars increases, demonstrating that retaining a small number of exemplars is still beneficial to stabilize the GNNs during training. Notably, OGCIL achieves high performance even with one exemplar. This is because a few exemplars in combination with knowledge distillation are enough to stabilize the representation space, thus enabling the generation of high-quality pseudo ID and OOD samples. Overall, OGCIL achieves competitive results consistently without relying on large exemplar sets.

\subsubsection{Exemplar Method and GNNs Backbones}

Additionally, we evaluate OGCIL (Figure~\ref{fig:mf}) by switching CM to the mean of features (MF) exemplar method~\cite{ergnn}, which selects nodes closest to the class prototype computed as the mean of their embeddings. Our results show that OGCIL maintains strong performance under both exemplar strategies. To further assess OGCIL’s adaptability, we swap the GCN backbone for GAT (Figure~\ref{fig:gat}), observing that OGCIL preserves its superiority with a different architecture. These findings suggest that OGCIL’s effectiveness is not tightly bound to a specific exemplar method or GNN backbone.

\subsection{t-SNE Visualization}

Figure~\ref{fig:tsne} visualizes node embeddings from OGCIL and baselines on the CS dataset using t-SNE. OGCIL shows A clear separation between known and unknown classes, with distinct clusters and minimal overlap, and effectively positions unknown samples (dark green points) away from known-class clusters. In contrast, OpenWGL and G$^2$Pxy embeddings show substantial overlap between known and unknown classes. ERGNN and EWC generate more compact known-class clusters but fail to isolate unknown classes effectively, whereas SSRM exhibits weaker known-class separation. The superior separation achieved by OGCIL highlights the effectiveness of its prototypical HSC loss and embedding-based pseudo-sample generation strategy.

\section{Conclusion}
In this paper, we present the first investigation into class-incremental open-set recognition within the graph domain to address the dual challenges of catastrophic forgetting and unknown class detection. To tackle these issues, we propose OGCIL, which generates pseudo samples in the embedding space using a prototypical conditional variational autoencoder, and refines class boundaries with a prototypical hypersphere classification loss. Extensive experimental results confirm the effectiveness of each component, demonstrating superior performance across five benchmark datasets in comparison to existing graph-based class-incremental and open-set methods.

\section{Appendix}
\subsection{Pseudo code for OGCIL}
Below is the pseudocode implementation. Our full code will be released on GitHub upon publication.
\begin{algorithm}[t]
\caption{OGCIL Training Procedure}
\label{alg:ogcil}
\SetKwInOut{Input}{Input}
\SetKwInOut{Output}{Output}

\Input{
  - A sequence of emerging graphs $\{\G^1,\dots,\G^M\}$ with disjoint class set $\{C^1,\dots,C^M\}$  \\
  - Model parameters $\Theta$: (GNN, $q_{\phi}(\cdot)$, $p_{\theta}(\cdot)$, $\mathbf{p}_c$, etc.) \\
  - Epochs $E$ per task, OOD regen interval $I$, $H$ and $L$  \\ as the number of pseudo ID and OOD samples \\
}
\Output{ 
  - Updated model parameters $\Theta$ 
}

\For{$t \leftarrow 1$ \KwTo $M$}{
    \uIf{$t = 1$}{
        
        \For{$\mathrm{epoch} \leftarrow 1$ \KwTo $E$}{
            \textbf{Train with} $ \mathcal{L}_{\mathrm{pcvae}} + \mathcal{L}_{\mathrm{phsc}}$ on $\mathcal{G}^t$ \\
            \quad -- Update $\Theta$ using gradient of these losses \\
            
        }
    }
    \Else{
        \textbf{Generate H ID embeddings} via $p_{\theta}(\cdot)$, $\{p_c\}_{c\in C^t}$ \\
        \textbf{Generate Examplar set $\mathcal{M}^t$} (e.g, CM)
        
        
        \For{$\mathrm{epoch} \leftarrow 1$ \KwTo $E$}{
            \If{$(\mathrm{epoch} \bmod I) = 0$}{
                \textbf{Generate L OOD embeddings} via mixing
            }
            \textbf{Train with} $\mathcal{L}_{\mathrm{phsc}} + \mathcal{L}_{\mathrm{pcvae}} + \lambda_{\mathrm{kd}} \mathcal{L}_{\mathrm{kd}}$ on $\mathcal{G}^t$, $\mathcal{M}^t$\\
            \quad -- Update $\Theta$ using gradient of combined losses
        }
    }
}
\end{algorithm}

\subsection{Datasets}

We evaluate our method on five datasets: Corafull~\cite{corafull}, Computer~\cite{computerphoto}, Photo~\cite{computerphoto}, CS~\cite{computerphoto}, and Arxiv~\cite{arxiv}. CoraFull is a citation network of research papers with bag-of-words features and topic labels. Following~\cite{twp}, we retain classes with over 150 nodes to ensure balanced data splits. Computers and Photos are Amazon co-purchase graphs, where nodes represent products with features from reviews, and edges indicate co-purchases. CS is a co-authorship network in computer science, with nodes representing authors and features derived from paper keywords. Arxiv is a citation network of CS papers with node features based on embeddings of titles and abstracts, along with publication year labels. 

Each dataset is partitioned into a sequence of tasks with disjoint classes, subdivided into known and unknown categories. To maximize data efficiency, we allow unknown classes in one task to become known in subsequent tasks. For instance, in CoraFull, the first task includes 8 known classes and 5 unknown classes. In the second task, 8 new known classes are introduced—5 of which were previously unknown, along with 5 entirely new unknown classes. This division not only enables the creation of more tasks but also mimics real-world scenarios where initially unseen classes emerge as known. Under the inductive setting, nodes from unknown classes are completely masked during training, ensuring the model has no prior exposure to unseen classes and maintains consistency with the principles of incremental learning. For data splits, we adopt the original test partition for Arxiv, while using a uniform 40\%/20\%/40\% split (training/validation/test) for all other datasets that do not provide a test partition. The characteristics of the datasets are summarized in Table~\ref{tab:data_statistics_pmgc}.
\begin{table}[t]
    \centering
    \caption{Statistics of datasets} 
    \vspace{-0.35cm}
    \resizebox{\linewidth}{!}{
    \begin{tabular}{cccccc}\\ \toprule
    \textbf{Dataset}  & \textbf{CoraFull} & \textbf{Computer}& \textbf{Photo}& \textbf{CS}& \textbf{Arxiv} \\ \hline     
    \# nodes & 19793 & 13752 & 7650 & 18333 & 169343 \\
    \# edges & 126842 & 491722 & 238162 & 163788 & 2315598 \\
    \# class & 45 & 10 & 8 & 15 & 40 \\ \hline
    \# tasks & 5 & 3 & 3 & 3 & 4\\
    \# knowns  & 8/8/8/8/8 & 4/2/2 & 3/2/2& 4/4/4 & 9/9/9/9 \\
    \# unknowns  & 5/5/5/5/5 & 2/2/2 & 1/1/1 & 3/3/3 & 4/4/4/4\\
    \# instance/class & 155 & 550 & 383 & 489 & 2274 \\
   \hline
    \end{tabular}}
    \label{tab:data_statistics_pmgc}
\end{table}

\subsection{Evaluation Metric}
We evaluate the performance of all methods using three metrics: open set classification rate (OSCR)~\cite{oscr}, close-set accuracy, and open-set AUC-ROC. OSCR is defined as the area under the CCR-FPR curve, where CCR (Correct Classification Rate) represents the proportion of correctly classified known samples above a threshold, and FPR (False Positive Rate) is the proportion of unknown samples classified as known. This metric captures the trade-off between correctly identifying known samples and minimizing misclassification of unknowns across varying thresholds. Close-set accuracy measures the model's classification performance when only known classes are present, serving as a baseline for inlier classification. AUC-ROC evaluates the model's ability to separate known and unknown classes by plotting true positive against false positive rates across thresholds. We choose OSCR and AUC-ROC because they ensure comparability across methods, as some baselines include explicit thresholding strategies while others do not, whereas these two metrics provide a threshold-agnostic way of performance evaluation. These metrics comprehensively evaluate class-incremental open-set performance.

\subsection{Implementation Details for OGCIL}

Our model is trained for up to 5000 epochs per task, with the best model stored based on the validation accuracy. The learning rate is set to 0.001. For the GCN backbone, we employ two hidden layers with a dimension of 256. Hyperparameter settings include $\lambda_{\mathrm{kd}} = 1$ for CS and $100$ for other datasets, and $\lambda_{\mathrm{reconst}} = 10$. The $\beta$ parameter for the Beta distribution is set to 5. To ensure computational efficiency, 100 OOD samples are generated and refreshed every 20 epochs. Additionally, 300 ID samples are sampled at the beginning of each task’s training phase. All results in the experimental sections are averaged over 5 independent runs.

\section*{Acknowledgements}

This work was supported in part by the National Natural Science Foundation of China under Grant 62262005, in part by the High-level Innovative Talents in Guizhou Province under Grant GCC[2023]033.

\bibliographystyle{ACM-Reference-Format}

\bibliography{mybib}

\end{document}